\PassOptionsToPackage{table}{xcolor}
\documentclass{fairmeta}

\usepackage{template/macro}

\usepackage{iftex}
\ifPDFTeX
  \usepackage[T1]{fontenc}
  \usepackage[utf8]{inputenc}
\fi
\usepackage[english]{babel}

\makeatletter
\@ifclassloaded{fairmeta}{}{%
  \usepackage{times}
}
\makeatother

\PassOptionsToPackage{hyphens}{url}
\usepackage{url}
\usepackage{hyperref}

\usepackage{graphicx}
\usepackage{caption}
\makeatletter
\@ifclassloaded{fairmeta}{}{%
  \captionsetup{margin=10pt,font=small,labelfont=bf}
}
\makeatother
\usepackage{tikz}
\usetikzlibrary{positioning,shapes.geometric,calc}
\usepackage{pgfplots}
\pgfplotsset{compat=1.17}

\usepackage{array}
\usepackage{booktabs}
\usepackage{microtype}
\usepackage{xcolor}
\usepackage{float}
\IfFileExists{enumitem.sty}{
  \usepackage{enumitem}
  \setitemize{itemsep=0pt,parsep=0pt,topsep=0pt}
  \setenumerate{itemsep=0pt,parsep=0pt,topsep=0pt}
}{}

\IfFileExists{algorithm.sty}{
  \usepackage{algorithm}
  \usepackage{algorithmicx}
  \usepackage{algpseudocode}
}{}

\usepackage{listings}
\lstdefinestyle{lean}{
  basicstyle=\ttfamily\small,
  breaklines=true,
  columns=fullflexible,
  keepspaces=true,
  xleftmargin=4pt,
  aboveskip=6pt,
  belowskip=6pt,
  literate=
    {ℝ}{{$\mathbb{R}$}}1
    {ℕ}{{$\mathbb{N}$}}1
    {→}{{$\to$}}1
    {←}{{$\leftarrow$}}1
    {∀}{{$\forall$}}1
    {∃}{{$\exists$}}1
    {≤}{{$\le$}}1
    {≥}{{$\ge$}}1
    {≠}{{$\ne$}}1
    {⊆}{{$\subseteq$}}1
    {∈}{{$\in$}}1
    {∉}{{$\notin$}}1
    {⁻¹}{{$^{-1}$}}1
    {∫}{{$\int$}}1
    {α}{{$\alpha$}}1
    {·}{{$\cdot$}}1
    {⟩}{{$\rangle$}}1
    {⟨}{{$\langle$}}1
}
\usepackage{natbib}

\usepackage[normalem]{ulem}

\usepackage[most]{tcolorbox}
\newtcolorbox{bookcardbox}[1][]{%
  enhanced,
  breakable,
  colback=green!5!white,
  colframe=green!45!black,
  colbacktitle=green!18!white,
  coltitle=green!30!black,
  fonttitle=\bfseries,
  title filled,
  arc=6pt,
  boxrule=0.6pt,
  left=8pt,right=8pt,top=5pt,bottom=5pt,
  toptitle=3pt,bottomtitle=3pt,
  #1
}
\makeatletter
\newcommand{\bookcard}[2]{%
  \def\bk@title{#1}\def\bk@author{#2}\bookcard@aux
}
\newcommand{\bookcard@aux}[8]{%
  \begin{bookcardbox}[title={\emph{\bk@title}, by \bk@author}]
    \footnotesize
    \setlength{\extrarowheight}{2pt}%
    {\sloppy\raggedright\href{#1}{\nolinkurl{#1}}\par}\smallskip
    \begin{tabular}{@{}p{0.30\linewidth}p{0.36\linewidth}p{0.30\linewidth}@{}}
      \textbf{Target statements:} #2 &
      \textbf{Formalized statements:} #5 &
      \\
      \textbf{Pages:} #3 &
      \textbf{Lines of code:} #4 &
      \textbf{Tokens (M):} #7 \\
    \end{tabular}\par\smallskip
  \end{bookcardbox}%
}
\makeatother

\usepackage{xspace}
\usepackage{nicefrac}
\usepackage{titletoc}

\ifdefined\pdfsuppresswarningpagegroup
\fi
\AtBeginDocument{\pagecolor{metapage}}
\let\cite\citep

\newtcolorbox{questionbox}{
  enhanced,
  frame hidden,
  colback=metabg,
  arc=7pt,
  boxrule=0pt,
  left=8pt,
  right=8pt,
  top=7pt,
  bottom=7pt,
  before skip=0.8em,
  after skip=0.8em,
}

\title{TMPO: Trajectory Matching Policy Optimization for Diverse and Efficient Diffusion Alignment}

\author[1,2,*]{Jiaming Li}
\author[1,*]{Chenyu Zhu}
\author[1]{Nanxi Yi}
\author[2]{Youjun Bao}
\author[2]{Li Sun}
\author[2]{Quanying Lv}
\author[3]{Xiang Fang}
\author[4]{Daizong Liu}
\author[1]{Jianjun Li}
\author[1]{Kun He}
\author[5]{Bowen Zhou}
\author[1,\dag]{Zhiyuan Ma}

\affiliation[1]{MAIR Lab, Huazhong University of Science and Technology}
\affiliation[2]{Kuaishou Technology}
\affiliation[3]{Nanyang Technological University}
\affiliation[4]{Wuhan University}
\affiliation[5]{Tsinghua University}

\contribution[*]{Equal contribution}
\contribution[\dag]{Corresponding author}

\abstract{Reinforcement learning (RL) has shown extraordinary potential in aligning diffusion models to downstream tasks, yet most of them still suffer from significant reward hacking, which degrades generative diversity and quality by inducing visual mode collapse and amplifying unreliable rewards. We identify the root cause as the \textit{mode-seeking} nature of these methods, which maximize expected reward without effectively constraining probability distribution over acceptable trajectories, causing concentration on a few high-reward paths. In contrast, we propose \textbf{Trajectory Matching Policy Optimization (TMPO)}, which replaces scalar reward maximization with trajectory-level reward distribution matching. Specifically, TMPO introduces a \textbf{Softmax Trajectory Balance (Softmax-TB)} objective to match the policy probabilities of $K$ trajectories to a reward-induced Boltzmann distribution. We prove that this objective inherits the \textit{mode-covering} property of forward KL divergence, preserving coverage over all acceptable trajectories while optimizing reward. To further reduce multi-trajectory training time on large-scale flow-matching models, TMPO incorporates \textbf{Dynamic Stochastic Tree Sampling}, where trajectories share denoising prefixes and branch at dynamically scheduled steps, reducing redundant computation while improving training effectiveness. Extensive results across diverse alignment tasks such as human preference, compositional generation and text rendering show that TMPO improves generative diversity over state-of-the-art methods by 9.1\%, and achieves competitive performance in all downstream and efficiency metrics, attaining the optimal trade-off between reward and diversity.
}

\date{\today}
\correspondence{Zhiyuan Ma at \email{mzyth@hust.edu.cn}}
\metadata[Code]{\url{https://github.com/MAIR-Lab-HUST/TMPO}}

\begin{document}

\maketitle
\begin{figure}[ht]
    \centering
    \includegraphics[width=\textwidth]{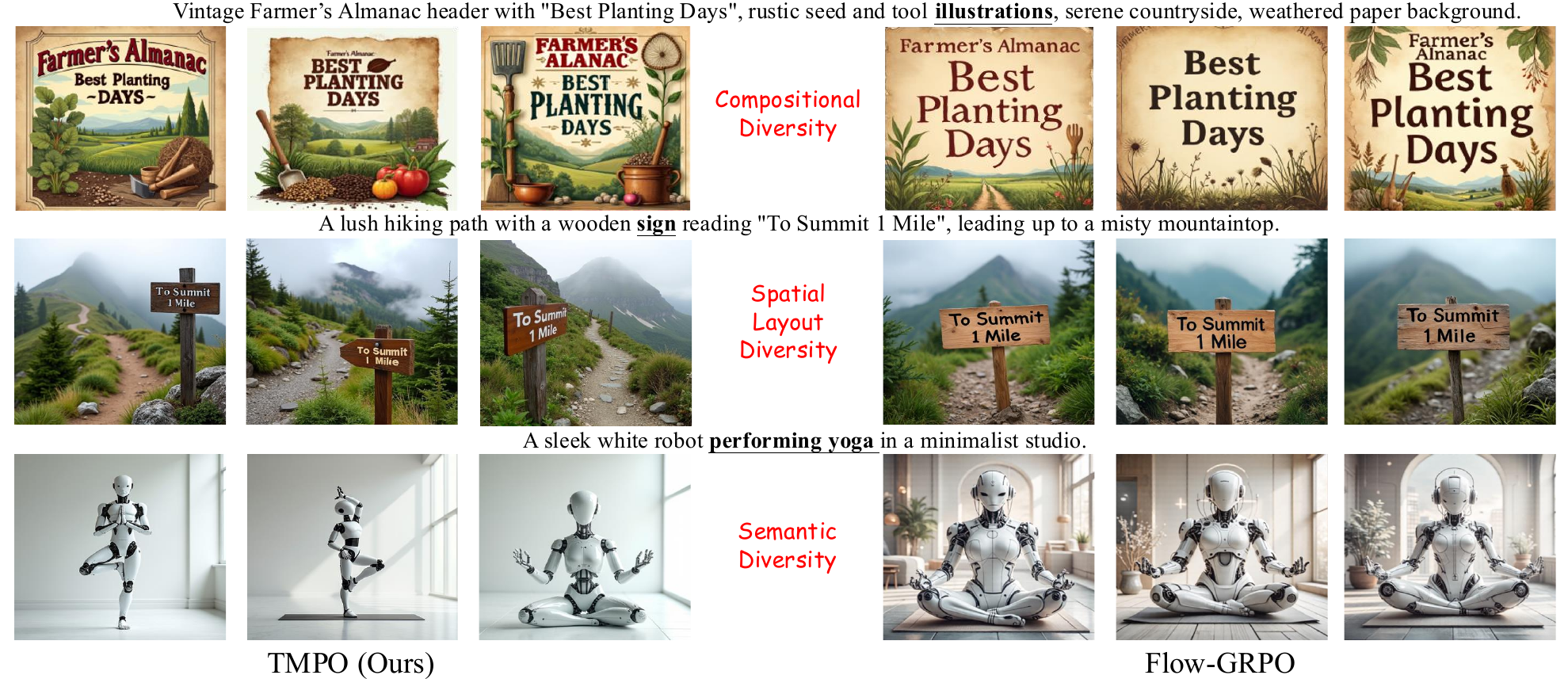}
    \caption{\textbf{Generative diversity comparison between TMPO (Ours) and Flow-GRPO.}}
    \label{fig:comparison}
\end{figure}

\section{Introduction}

\begin{figure}[ht]
    \centering
    \includegraphics[width=\textwidth]{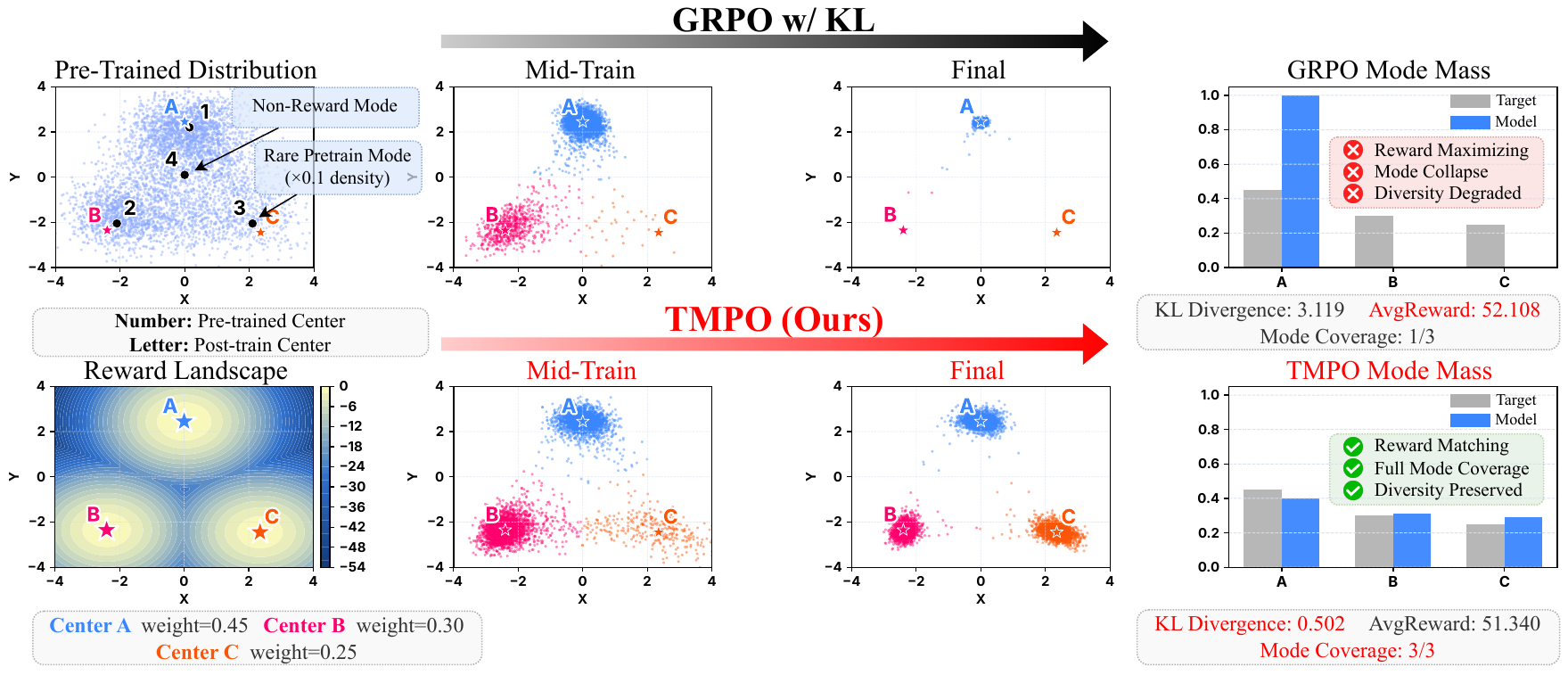}
    \caption{\textbf{A toy experiment on a three-layer MLP diffusion model pre-trained on a Gaussian mixture distribution containing non-reward modes and rare density modes.} The post-training objective is to match a multimodal reward distribution. GRPO-based reward maximization collapses to a single high-reward mode even with KL regularization, whereas TMPO covers all reward clusters by matching trajectory-level reward distributions while maintaining a high average reward.}
    \label{fig:toy_experiment}
\end{figure}

Recently, RL has achieved remarkable success in aligning Large Language Models (LLMs), generative diffusion models and flow-matching models. However, existing methods still suffer from significant \textit{reward hacking} issue, which implies the model may ignore the distribution diversity and instead take a shortcut by over-reinforcing a small proportion of the recognized high-reward modes. Similar to LLM, this dilemma can be called \textit{visual mode collapse}. While mitigating reward hacking has been extensively studied in the field of LLMs~\cite{moskovitz2023confronting,coste2023reward}, the issue remains largely unexplored for diffusion or flow models, which raises two critical issues: \emph{\textbf{1) Degradation of Diversity:}} for language models, a problem may solely own a single definite answer, whereas for generative models (\emph{i.e.,} text-to-image models), the same prompt may correspond to many plausible outputs that vary in \emph{composition}, \emph{spatial layout}, and \emph{scene semantics}. Reward hacking, however, drives the model to generate images with only a small set of compositions and styles, degrading generative diversity. (Fig.~\ref{fig:comparison}). \emph{\textbf{2) Amplification of Unreliable Rewards:}} for text-to-image models, both model-based rewards (\emph{e.g.,} ImageReward~\cite{xu2023imagereward}, HPS~\cite{wu2023hpsv2}, PickScore~\cite{kirstain2023pick}) and rule-based rewards (\emph{e.g.,} OCR accuracy and GenEval~\cite{ghosh2024geneval}) are intrinsically imperfect proxies for human preferences~\cite{gao2023scaling,chen2023textdiffuser,pan2022the,skalse2022defining,he2025gardo}. Reward hacking on these proxy rewards amplifies spuriously rewarded image attributes, thereby deteriorating generation quality. This raises a core question in generative diffusion alignment: \emph{How can we preserve the anisotropic distribution of outputs learned by the model from pre-training data while encouraging the model to satisfy rewards with specific preferences?}


The reward hacking behavior in diffusion model alignment can be attributed to the training objectives of current diffusion-based RL methods. Existing methods use reward maximization as the optimization objective~\cite{DDPO,fan2023dpok,liu2025flowgrpo,xue2025dancegrpo,li2025mixgrpo,ding2026treegrpo}, which we prove is intrinsically \textit{mode-seeking}. In particular, mode-seeking models tend to explore high-reward trajectory, rather than covering all reasonable counterparts, as illustrated by the GRPO example in Fig.~\ref{fig:toy_experiment}. \textbf{An ideal optimization objective should be \textit{mode-covering}, which encourages the model to cover as many reasonable solutions as possible.} This analysis further suggests that alleviating reward hacking requires a fundamental shift in the optimization objective. Existing methods mainly attempt to address the issue of reward hacking through two lines of work: \emph{\textbf{1) Constraining Over-optimization:}} KL regularization~\cite{ouyang2022training,stiennon2020learning,fan2023dpok,liu2025flowgrpo,he2025gardo} is used to constrain the deviation of the policy from the reference model, but it does not alter the reward-maximization objective, resulting in limited improvement in generative diversity. \emph{\textbf{2) Improving Training Objective:}} Generative Flow Network (GFlowNet)-based methods~\cite{zhang2024dgfs,zhang2025improving,liu2025nablagfn} seek to generate outputs with probabilities proportional to their rewards, but existing methods usually establish local constraints on individual denoising steps, which requires estimating an intractable state-flow function for policy updates, thereby introducing additional training error. Overall, existing mitigation methods still fall short of providing a direct mode-covering objective over the full denoising trajectory.

To address this gap, we propose \textbf{Trajectory Matching Policy Optimization (TMPO)}. Inspired by the reward distribution matching nature of GFlowNets~\cite{bengio2021flow}, TMPO fundamentally shifts the training objective from reward maximization to trajectory-level reward distribution matching. We prove that TMPO is \textit{mode-covering} and can naturally encourage the coverage of reasonable solutions, thereby alleviating reward hacking from the ground up, as shown in Fig.~\ref{fig:toy_experiment}. Specifically, we introduce a \textbf{Softmax Trajectory Balance (Softmax-TB)} objective, which performs reward distribution matching on the complete denoising trajectory. For $K$ trajectories generated from the same starting point, Softmax-TB matches the generation probabilities of these trajectories to a target Boltzmann distribution computed from rewards ( $P_\theta(\tau) \propto \exp(\beta R(\tau))$ ). Meanwhile, by normalizing the trajectory probabilities $P_\theta(\tau)$ and exponential rewards $\exp(\beta R(\tau))$ among the $K$ trajectories within each group, we eliminate the intractable partition function, yielding a simpler and directly optimizable objective. To reduce the training time of fully sampling multiple trajectories under Softmax-TB~\cite{zhang2025improving}, TMPO further introduces \textbf{Dynamic Stochastic Tree Sampling}~\cite{li2025mixgrpo,fu2025dynamictreerpo,ding2026treegrpo}, which allows trajectories to share denoising steps through a tree structure, reducing redundant computation and lowering training time. By dynamically adjusting the branching points, the tree sampler also promotes effective exploration at different denoising stages and improves training performance.

\begin{figure}[h]
    \centering
    \includegraphics[width=0.48\linewidth]{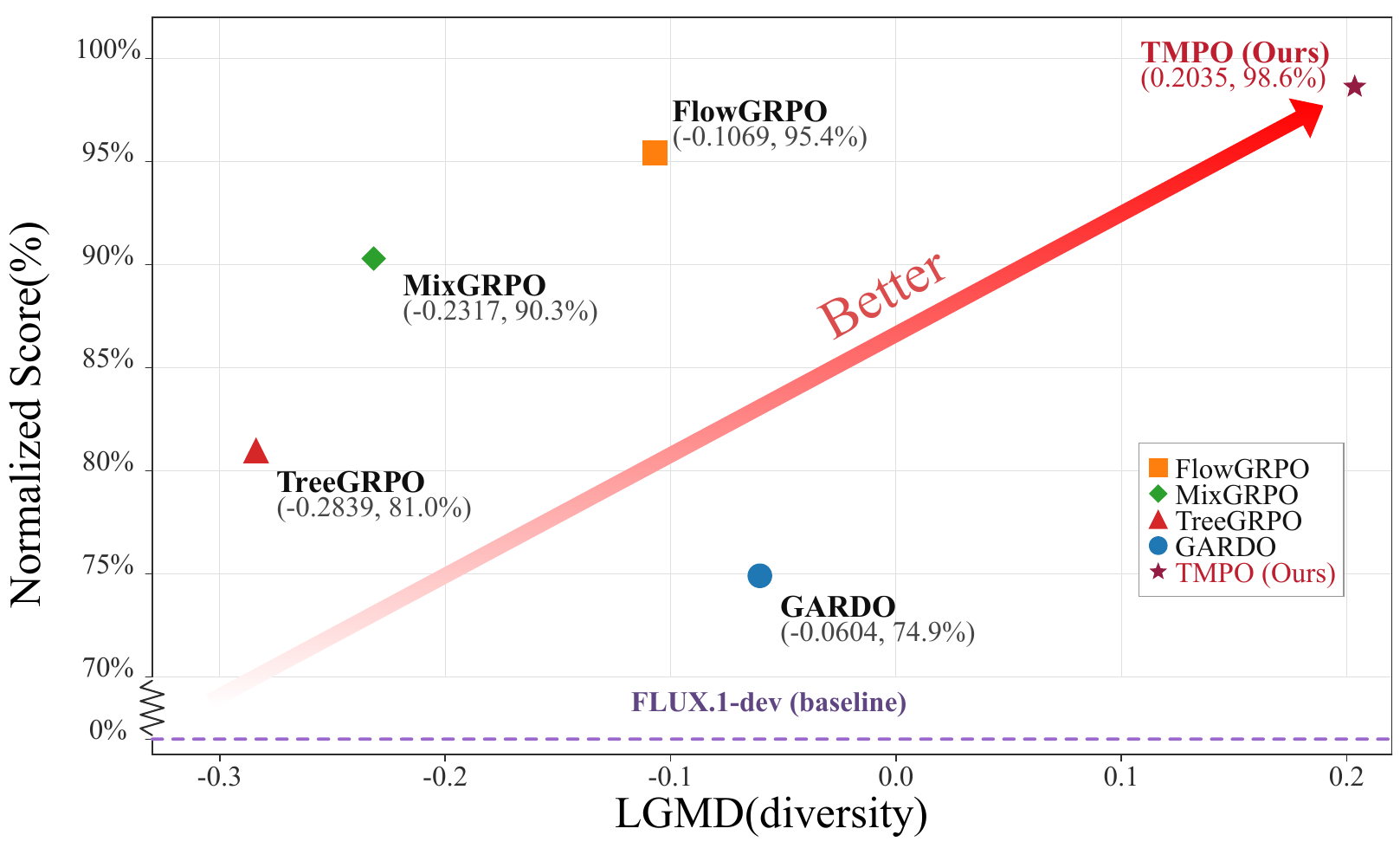}
    \hfill
    \includegraphics[width=0.48\linewidth]{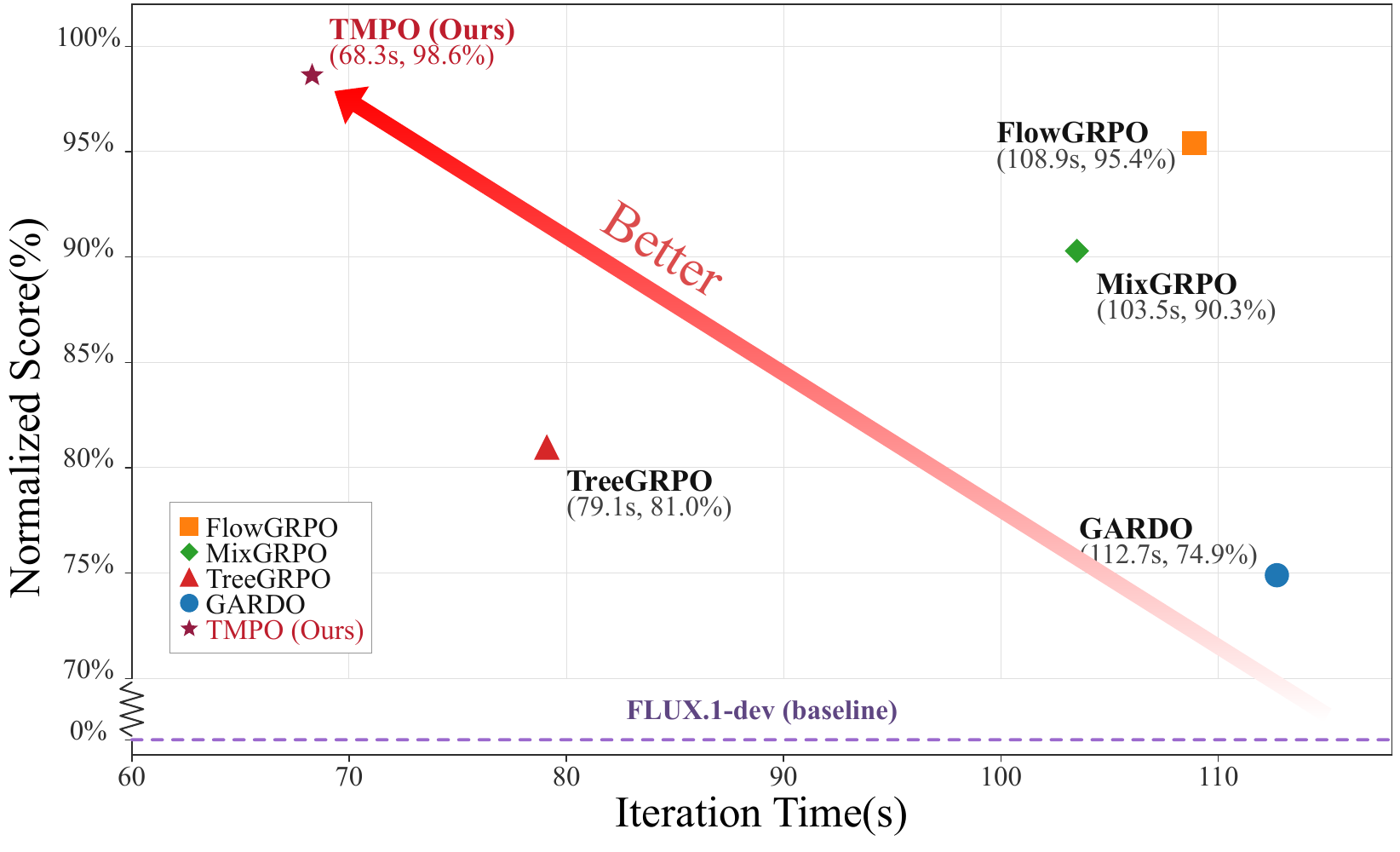}
    \caption{\textbf{Pareto analysis of TMPO against GRPO-based alignment methods.} \textbf{Left:} normalized reward score versus generation diversity measured by LGMD. \textbf{Right:} normalized reward score versus iteration time. Normalized scores are computed via min-max normalization across all compared methods for each metric. TMPO achieves the most favorable trade-off in both reward-diversity and reward-efficiency comparisons.}
    \label{fig:pareto}
\end{figure}

We evaluate TMPO on FLUX.1-dev across three alignment tasks: compositional image generation, visual text rendering, and human preference alignment. Results show that TMPO obtains the highest diversity metrics across all tasks, improving generative diversity by 9.1\% on average over existing state-of-the-art methods, while also achieving the best GenEval accuracy and PickScore. Moreover, TMPO reduces training time by up to 27\% compared with state-of-the-art methods and achieves the most favorable trade-offs in both reward-diversity and reward-efficiency comparisons, as shown in Fig.~\ref{fig:pareto}.

Our contributions are as follows: \textbf{(1)} We identify reward maximization as the root cause of reward hacking in diffusion-based RL and reformulate the optimization objective as reward distribution matching. \textbf{(2)} We introduce \textbf{Softmax Trajectory Balance}, a partition-function-free trajectory-level distribution matching objective, and prove that its mode-covering property naturally preserves generative diversity. \textbf{(3)} We introduce \textbf{Dynamic Stochastic Tree Sampling}, which substantially reduces the training time of our trajectory balance objective and enables scalable training on large-scale flow-matching models.

\section{Related Work}

\paragraph{Mitigating reward hacking in diffusion model alignment.}
Diffusion-based RL alignment is highly susceptible to reward hacking under model-based or rule-based rewards~\cite{pan2022the,skalse2022defining,hong2026understanding,wu2025rewarddance}. Existing mitigations, including KL regularization~\cite{fan2023dpok,liu2025flowgrpo}, pairwise preference rewards~\cite{wang2025pref}, and diversity-aware regularization~\cite{he2025gardo}, still optimize scalar rewards and thus inherit mode-collapse risks~\cite{gx2025kl}. TMPO addresses this limitation by reformulating scalar rewards as a preference distribution to be matched.

\paragraph{Distribution matching and GFlowNet-based alignment.}
GFlowNets train stochastic policies with terminal distributions proportional to rewards~\cite{bengio2021flow}. DGFS~\cite{zhang2024dgfs}, DAG~\cite{zhang2025improving}, and $\nabla$-GFlowNet~\cite{liu2025nablagfn} adapt this idea to diffusion alignment using local detailed-balance objectives, with $\nabla$-GFlowNet further incorporating reward gradients. However, their local or partial-trajectory constraints require intractable state-flow estimation and can accumulate errors over long denoising horizons. TMPO instead directly matches complete trajectory probabilities to a reward-induced Boltzmann distribution, avoiding per-transition flow estimation and explicit partition computation.

\paragraph{Efficient trajectory sampling in diffusion model alignment.}
RL post-training for large-scale flow models is bottlenecked by redundant rollouts, and complete-trajectory objectives amplify this cost by requiring policy probabilities over multiple trajectories~\cite{zhang2025improving}. MixGRPO reduces stochastic optimization to selected SDE windows~\cite{li2025mixgrpo}, while Dynamic-TreeRPO and TreeGRPO amortize early computation by sharing denoising prefixes~\cite{fu2025dynamictreerpo,ding2026treegrpo}. TMPO adopts prefix sharing and further introduces dynamic branching schedules to maintain effective exploration across training stages.

\section{Preliminaries}
\label{gen_inst}

\subsection{Trajectory Balance and Reward Matching}

Generative Flow Networks (GFlowNets) train stochastic policies to sample terminal objects in proportion to non-negative rewards~\cite{bengio2021flow}. For a trajectory $\tau=(s_0\to\cdots\to s_n)$ ending at $x=s_n$ with reward $R(x)$, Trajectory Balance (TB) imposes:~\cite{malkin2022trajectory}

\begin{equation}
    Z_{\theta}\prod_{t=1}^{n}P_F(s_t\mid s_{t-1};\theta)
    =
    R(x)\prod_{t=1}^{n}P_B(s_{t-1}\mid s_t;\theta),
    \label{eq:prelim_tb}
\end{equation}
where $P_F$, $P_B$, and $Z_\theta$ are the forward policy, backward policy, and total flow. This reward-proportional view preserves multiple high-reward modes. Diffusion GFlowNet methods and Flow-GRPO similarly motivate replacing scalar reward maximization with reward distribution matching~\cite{zhang2024dgfs,liu2025nablagfn,zhu2025flowrl}.

\subsection{ODE-to-SDE Conversion for Stochastic Diffusion Policies}

Modern text-to-image generators are commonly built on diffusion or flow-matching models~\cite{ho2020denoising,rombach2022high,lipman2023flow,liu2023flow,albergo2023stochastic,esser2024scaling,flux2024}. In rectified flow, a clean sample $x_0$ and noise $x_1\sim\mathcal{N}(0,\mathbf{I})$ are connected by:
\begin{equation}
    x_t=(1-t)x_0+t x_1,\qquad
    d x_t=v_{\theta}(x_t,t,c)\,dt ,
    \label{eq:prelim_rf}
\end{equation}
Deterministic ODE solvers~\cite{song2021denoising,lu2022dpmsolver} lack the transition probabilities required by policy-gradient RL. Following prior work~\cite{song2020score,liu2025flowgrpo,xue2025dancegrpo}, we convert the probability-flow ODE to an equivalent SDE that admits tractable likelihoods for credit assignment while preserving marginals:
\begin{equation}
    d x_t =
    \left[
        v_{\theta}(x_t,t,c)
        + \frac{1}{2}\sigma^2(t)\nabla_{x}\log p_{\theta}(x_t\mid c,t)
    \right]dt
    + \sigma(t)dW_t .
    \label{eq:prelim_probability_flow_sde}
\end{equation}
Here $\sigma(t)$ controls the noise scale, with $\sigma(t)\equiv 0$ recovering the deterministic ODE.

\section{Trajectory Matching Policy Optimization}
\label{headings}

\begin{figure*}[ht]
  \centering
  \includegraphics[width=\textwidth]{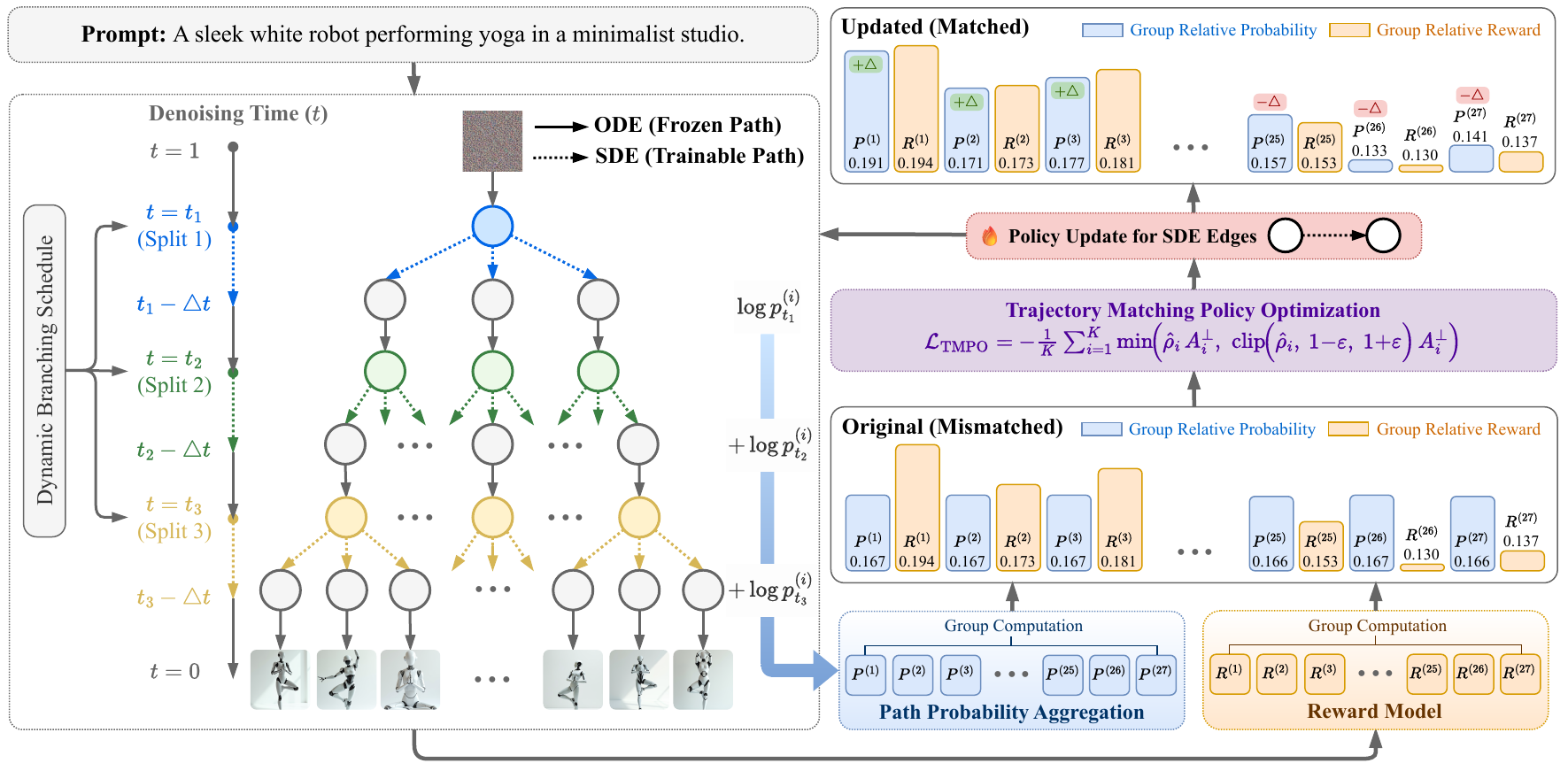}
  \caption{\textbf{Overview of the TMPO framework.} For each prompt, Dynamic Stochastic Tree Sampling shares a deterministic ODE prefix and injects SDE noise at three curriculum-scheduled branch points, yielding $3^3{=}27$ trajectories with reduced compute. Terminal images are scored by a reward model, and Softmax Trajectory Balance converts the rewards into a Boltzmann target, computing the per-trajectory advantage as the log-ratio between the target and policy distributions.}
  \label{fig:pipeline}
  \vspace{-1.5em}
\end{figure*}

In this section, we present \textbf{Trajectory Matching Policy Optimization (TMPO)} for online reward-induced trajectory distribution matching in flow-matching text-to-image models. We adapt GFlowNet trajectory balance to within-prompt trajectory groups, yielding a partition-free \textbf{Softmax-TB} objective with provable mode-covering properties (Section~\ref{prop:mode_covering}). We then describe how reward-bearing samples are efficiently collected through prefix-sharing SDE tree rollouts with \textbf{Dynamic Stochastic Tree Sampling}. The overall pipeline is illustrated in Figure~\ref{fig:pipeline}.

\subsection{From GFlowNets to Softmax Trajectory Balance}

GFlowNets learn a policy $\pi_\theta$ such that $P_\theta(\tau) \propto R(\tau)$ by minimizing a trajectory balance (TB) residual:
\begin{equation}
    \mathcal{L}_{\text{TB}}(\tau) = \bigl( \log Z + \log P_{\theta}(\tau) - \log R(\tau) \bigr)^2,
\end{equation}
where $Z$ is the partition function typically estimated via an auxiliary network. In TMPO's tree topology, all $K$ trajectories for the same prompt share a common $Z$, so normalizing within the group cancels the partition function exactly: $P_\theta(\tau_i)/\sum_j P_\theta(\tau_j) = R(\tau_i)/\sum_j R(\tau_j)$ (Appendix~\ref{appendix:theoretical_derivations}).

To introduce adjustable sharpness, we replace raw rewards with a Boltzmann target $p^*(\tau) \propto \exp(\beta R(\tau))$, where $\beta$ controls mode concentration. This is the maximum-entropy distribution subject to the expected reward constraint (Appendix~\ref{appendix:theoretical_derivations}). Substituting into the within-group matching and defining shorthand $q_i \triangleq \mathrm{softmax}_i(\beta R)$ and $p_i \triangleq P_\theta(\tau_i)/\sum_j P_\theta(\tau_j)$, we obtain the \textbf{Softmax-TB advantage}:
\begin{equation}
    A_i = \log q_i - \log p_i = \log \frac{\exp(\beta R_i)}{\sum_{j=1}^K \exp(\beta R_j)} - \log \frac{P_{\theta}(\tau_i)}{\sum_{j=1}^K P_{\theta}(\tau_j)}
\end{equation}
The following result shows that the Softmax-TB advantage is intimately connected to forward KL minimization.

\paragraph{Mode-Covering Equilibrium.}\label{prop:mode_covering}
Let $q_i = \exp(\beta R_i)/\sum_{j=1}^K \exp(\beta R_j)$ denote the within-group Boltzmann target and $p_i = P_\theta(\tau_i)/\sum_{j=1}^K P_\theta(\tau_j)$ the within-group policy distribution. Then the Boltzmann-weighted Softmax-TB advantage equals the within-group forward KL divergence:
\[
\sum_{i=1}^K q_i \, A_i = D_{\mathrm{KL}}^{(K)}(q_\beta \,\|\, p_\theta) \;\geq\; 0,
\]
with equality if and only if $p_i = q_i$ for all $i$. Because $D_{\mathrm{KL}}^{(K)}(q \| p) \to +\infty$ whenever $p_i \to 0$ for any $i$ with $q_i > 0$, Softmax-TB inherently penalizes under-coverage of any mode that carries positive Boltzmann weight. Within-group normalization cancels $Z_\beta$ and $Z_\theta$, making this conditional forward KL exactly computable over the $K$ observed trajectories, without requiring independent samples from the global intractable $q_\beta$ (extended proof and quantitative bounds in Appendix~\ref{appendix:forward_kl}).

\textit{Proof.} By definition, $A_i = \log q_i - \log p_i$. Substituting into the Boltzmann-weighted sum: $\sum_i q_i A_i = \sum_i q_i (\log q_i - \log p_i) = D_{\mathrm{KL}}^{(K)}(q \| p)$. Non-negativity and the equality condition follow from Gibbs' inequality. Since both $q$ and $p$ are normalized within the group, the global partition function $Z_\beta$ and the policy's total probability $Z_\theta$ cancel and need not be estimated. \hfill$\square$

\paragraph{Reverse KL vs.\ Forward KL.}\label{remark:mode_seeking}
Standard policy gradient methods sample from $\pi_\theta$ and therefore exhibit reverse-KL-like, mode-seeking behavior: under the entropy-regularized Boltzmann formulation, reward maximization (e.g., Flow-GRPO, TreeGRPO) can be interpreted as minimizing $D_{\mathrm{KL}}(p_\theta \| q_\beta)$ (Appendix~\ref{appendix:reverse_kl_comparison}). Its advantage $A_i \propto R_i - \bar{R}$ is agnostic to the policy probability $p_i$, so dropping a mode incurs no direct gradient penalty. In contrast, TMPO's log-ratio advantage $A_i = \log(q_i/p_i)$ is distribution-aware: it diverges whenever any mode with positive Boltzmann weight is under-covered ($p_i \to 0 \Rightarrow A_i \to +\infty$), inheriting the mode-covering behavior of forward KL despite sampling from the policy. \textbf{This distinction---the distribution-aware advantage of Equation~\eqref{eq:advantage}---provides the theoretical basis for diversity preservation without auxiliary KL regularization.}

\paragraph{Complete Softmax-TB Objective}
Since $A_i$ involves the shared softmax denominator that couples all $K$ trajectories, we detach it as a stop-gradient signal and route the gradient exclusively through the IS ratio $\hat{\rho}_i$. The complete TMPO loss is:
\begin{gather}
\boxed{
\mathcal{L}_{\text{TMPO}} = -\frac{1}{K}\sum_{i=1}^{K} \min\!\Bigl(\hat{\rho}_i \, A_i^{\bot},\;\operatorname{clip}\!\bigl(\hat{\rho}_i,\;1{-}\varepsilon,\;1{+}\varepsilon\bigr)\, A_i^{\bot}\Bigr)
} \label{eq:softmax_tb_loss} \\[4pt]
A_i = \log\frac{\exp(\beta R_i)}{\sum_{j=1}^K\exp(\beta R_j)} - \log\frac{P_\theta(\tau_i)}{\sum_{j=1}^K P_\theta(\tau_j)},\qquad \hat{\rho}_i = \prod_{t=1}^{T}\frac{\pi_\theta(x_{s_t^-}^{(i)} \mid x_{s_t}^{(i)})}{\pi_{\theta_{\text{old}}}(x_{s_t^-}^{(i)} \mid x_{s_t}^{(i)})} \label{eq:advantage}
\end{gather}
Here $A_i^{\bot}$ denotes the stop-gradient advantage and $\hat{\rho}_i$ is clipped to $[1{-}\varepsilon, 1{+}\varepsilon]$ for trust-region protection~\cite{schulman2017proximal}. The per-step log-ratios are centered via RatioNorm~\cite{wang2025grpo} to remove the deterministic negative shift of Gaussian transition kernels and restore symmetric clipping (Appendix~\ref{appendix:gradient_analysis}).

\subsection{Dynamic Stochastic Tree Sampling}
\label{sec:tree_sampling}

Unlike the independent parallel rollouts in standard GRPO-based methods~\cite{liu2025flowgrpo,xue2025dancegrpo}, TMPO places three consecutive stochastic branch points along the denoising trajectory, producing a $3^3{=}27$ terminal tree per prompt. At each branch point, $B{=}3$ child trajectories are spawned from independent noise realizations. Early branch points at high noise levels introduce coarse-grained diversity across trajectory groups, while later branch points at lower noise levels supply finer-grained structured variations. Shared prefix computations before each bifurcation substantially reduce the cost of RL training on large-scale models such as FLUX.

\paragraph{Dynamic Branching Schedule}
Branch positions follow a curriculum that shifts from early high-noise regions toward later structured regions as training progresses ($p = \mathrm{clip}(u/U, 0, 1)$). To prevent overfitting to a fixed tree geometry, each position is stochastically perturbed:
\begin{equation}
\xi_i \sim \mathrm{Beta}(\kappa \bar{\mu}_i(p),\, \kappa(1{-}\bar{\mu}_i(p))), \quad \tilde{s}_i = \lfloor s_{\min} + (s_{\max} - s_{\min})\xi_i + 0.5 \rfloor
\end{equation}
with final indices obtained by sorting. At each branch point $s_i$, child nodes are generated by injecting SDE noise with magnitude $\gamma(\sigma) = \eta\sqrt{\sigma/(1{-}\sigma)}\sqrt{-\Delta t}$, where $\eta{=}0.7$ following the CPS recommendation~\cite{wang2025cps}:
\begin{equation}
x_{s_i^-}^{(r)} = \mu_{\theta}(x_{s_i}, s_i) + \gamma(\sigma_{s_i})\varepsilon_r, \quad \varepsilon_r \sim \mathcal{N}(0, \mathbf{I})
\end{equation}
Between branch points, the denoising trajectory follows deterministic ODE steps that do not require gradient computation. The trajectory log-probability accumulates only the $T$ stochastic transitions at branch points: $\log P_\theta(\tau) = \sum_{i=1}^{T} \log \pi_\theta(x_{s_i^-} \mid x_{s_i})$, so gradient back-propagation is confined to $T$ steps rather than the full denoising horizon, substantially reducing per-iteration training cost. Full derivations of the SDE noise injection, Beta branching schedule, and log-probability computation are provided in Appendix~\ref{appendix:tree_sampling}.

\section{Experiments}
\label{sec:experiments}

This section empirically evaluates TMPO on three tasks: (1) Compositional Image Generation, (2) Visual Text Rendering, and (3) Human Preference Alignment. Beyond alignment quality, we also assess generative diversity and per-iteration training efficiency.

\subsection{Experimental setup}

\paragraph{Models, prompts, and metrics.}
We use FLUX.1-dev as the backbone, fine-tuned with LoRA~\cite{hu2022lora} (rank $r{=}64$, $\alpha{=}128$) targeting all attention and feed-forward projections in each transformer block. All images are generated at $512{\times}512$ resolution. Training rollouts use $6$ denoising steps for efficiency; all evaluations use $28$ steps. For human preference alignment, training and evaluation prompts are drawn from HPDv2~\cite{wu2023hpsv2}. For compositional image generation (GenEval) and visual text rendering (OCR), we construct task-specific training and test prompt sets. All evaluations generate 10 images per prompt. Task-specific metrics include GenEval~\cite{ghosh2024geneval} accuracy for composition, OCR accuracy ($1{-}\text{NED}$) for text rendering, and HPS-v2.1, ImageReward, PickScore for preference alignment. Diversity is measured by two complementary metrics. \textbf{Log Geometric Mean Distance (LGMD)} computes the log geometric mean of pairwise distances in VAE~\cite{kingma2014auto} latent space: $\text{LGMD} = \frac{2}{N(N{-}1)} \sum_{i<j} \log \bigl(\|\phi(x_i) - \phi(x_j)\|_2 / \sqrt{D}\bigr)$, where $\phi(\cdot)$ is the flattened VAE latent and $D$ the feature dimension; positive values indicate healthy diversity, while negative values signal mode collapse. \textbf{Cosine Diversity (Cos.\,Div.)} follows GARDO~\cite{he2025gardo} and measures the mean pairwise cosine distance in DINOv2~\cite{oquab2024dinov2} ViT-L/14 feature space: $\text{Cos.\,Div.} = \frac{2}{N(N{-}1)}\sum_{i<j}(1 - \cos(\psi(x_i), \psi(x_j)))$. LGMD captures low-level structural duplicates, while Cos.\,Div.\ captures semantic layout and texture differences.

\paragraph{Training protocols and baselines.}
We consider three single-reward protocols, namely GenEval only (compositional accuracy), OCR only (text rendering accuracy), and PickScore~\cite{kirstain2023pick} only (human preference), as well as joint preference training with HPS-v2.1, ImageReward, and PickScore at equal weight, where each reward is independently z-score normalized within the $K$-trajectory group before summation. TMPO uses a three-level prefix-sharing tree with $K{=}3^3{=}27$ terminal trajectories per prompt. TMPO is compared against Flow-GRPO~\cite{liu2025flowgrpo}, MixGRPO~\cite{li2025mixgrpo}, TreeGRPO~\cite{ding2026treegrpo}, and GARDO~\cite{he2025gardo} under matched backbone and reward settings. All GRPO-based baselines adopt KL regularization with $\beta_{\text{KL}}{=}0.03$. Full hyperparameter specifications, model details, and compute resources are provided in Appendix~\ref{app:sec:exp-setup-details}.

\subsection{Main results}

Table~\ref{tab:main_results} summarizes results across three training protocols. Joint multi-reward results are provided in Table~\ref{tab:comparison_joint} (Appendix~\ref{app:sec:extended-exp-results}). TMPO also generalizes to SD3.5-Medium, outperforming all baselines under the PickScore only protocol (Table~\ref{tab:sd35} in Appendix~\ref{app:subsec:sd35}).

\begin{figure*}[h]
    \centering
    \begin{subfigure}[t]{0.32\textwidth}
        \centering
        \includegraphics[width=\linewidth]{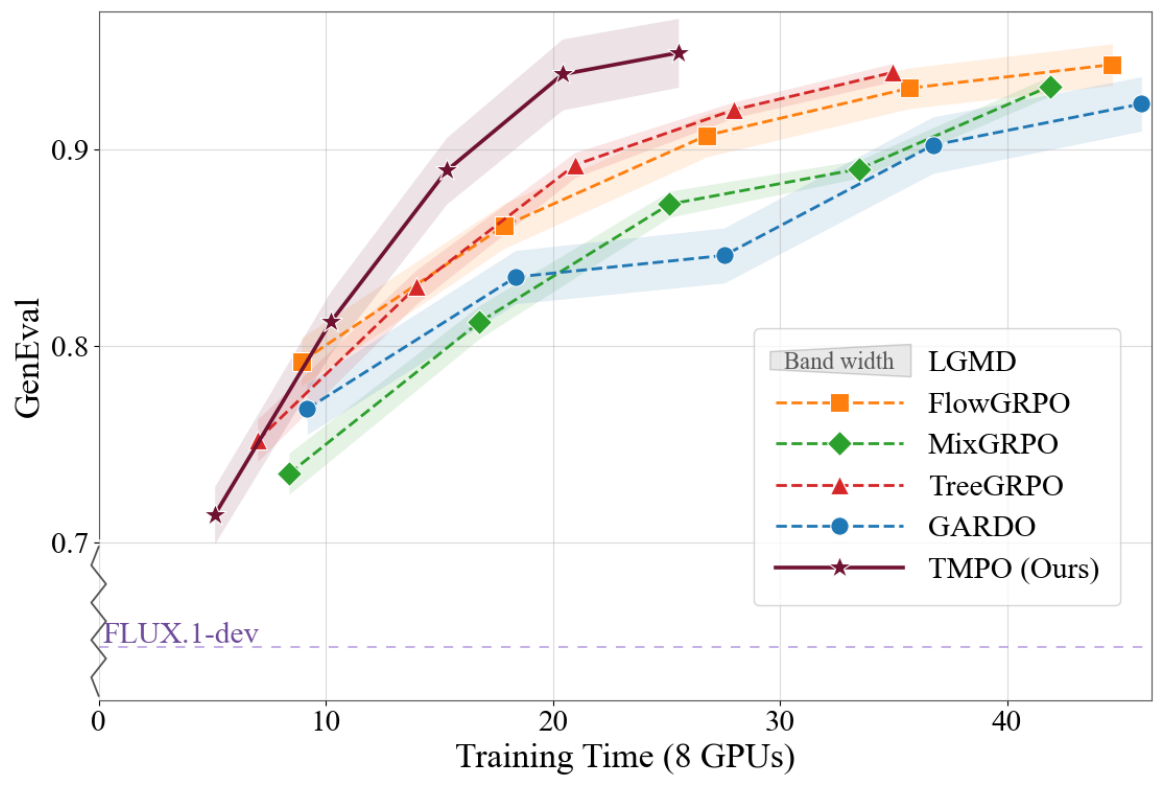}
        \caption{\textbf{GenEval (composition)}}
        \label{fig:training_step_geneval}
    \end{subfigure}
    \hfill
    \begin{subfigure}[t]{0.32\textwidth}
        \centering
        \includegraphics[width=\linewidth]{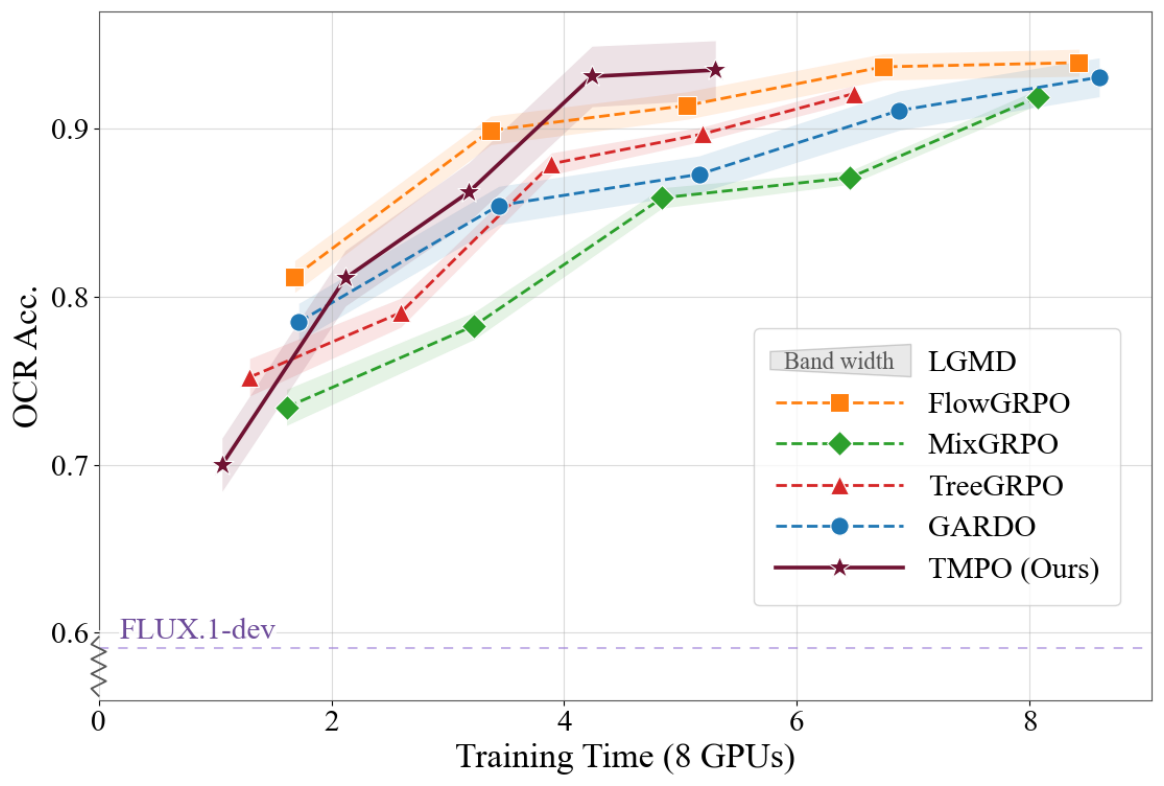}
        \caption{\textbf{OCR Acc.\ (text rendering)}}
        \label{fig:training_step_ocr}
    \end{subfigure}
    \hfill
    \begin{subfigure}[t]{0.32\textwidth}
        \centering
        \includegraphics[width=\linewidth]{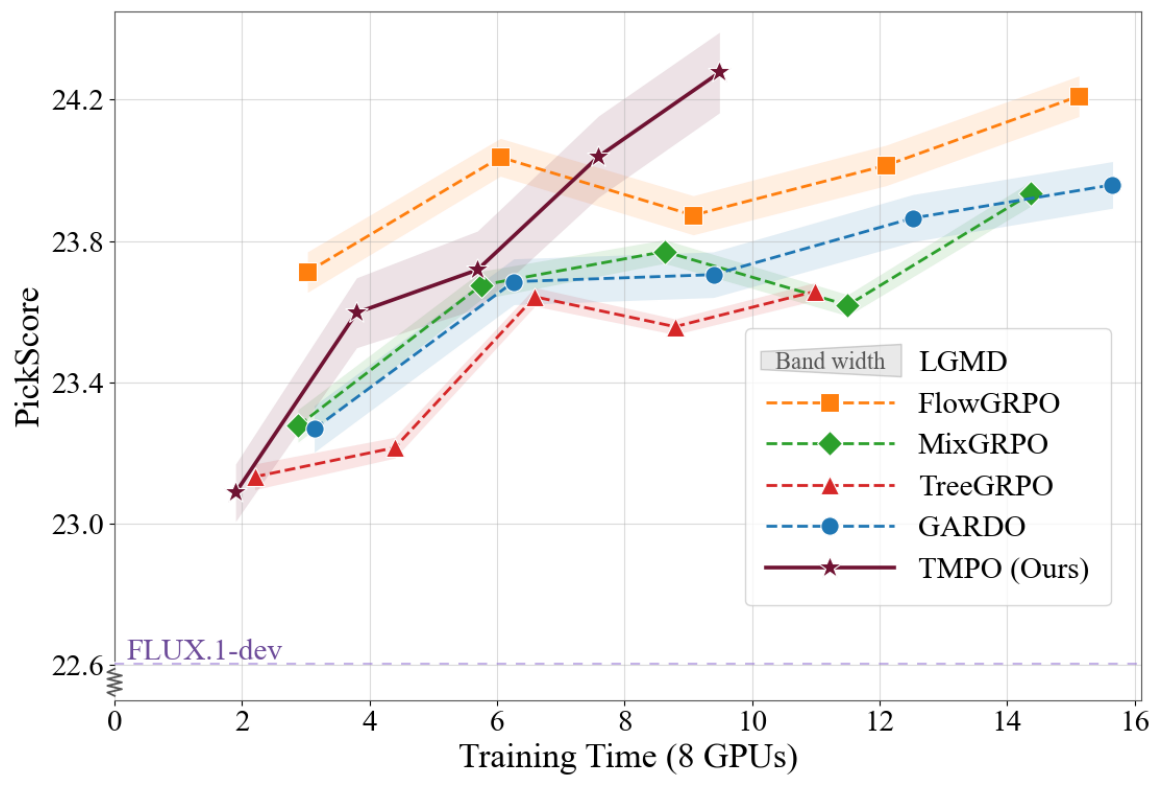}
        \caption{\textbf{PickScore (preference)}}
        \label{fig:training_steps_pickscore}
    \end{subfigure}
    \caption{\textbf{Training curves across three single-reward protocols.} Each plot shows the task reward versus wall-clock training time. TMPO converges faster and reaches higher final reward than all baselines, benefiting from both prefix-sharing tree sampling and Softmax-TB distribution matching.}
    \label{fig:training_curves}
\end{figure*}

\paragraph{Compositional image generation.}
Under GenEval only training, TMPO achieves the best compositional accuracy (0.949) while preserving superior diversity on both LGMD and Cosine Diversity. Despite the sparse binary reward signal, Softmax-TB effectively distributes probability mass toward compositionally correct trajectories without collapsing onto a narrow set of solutions.

\paragraph{Visual text rendering.}
TMPO achieves the best iteration time, the highest HPS-v2.1 and PickScore, and the best diversity on both LGMD and Cosine Diversity. Although Flow-GRPO reaches the highest OCR accuracy, it substantially reduces diversity, indicating stronger overfitting to the task reward.

\begin{figure*}[h]
  \centering
  \includegraphics[width=\textwidth]{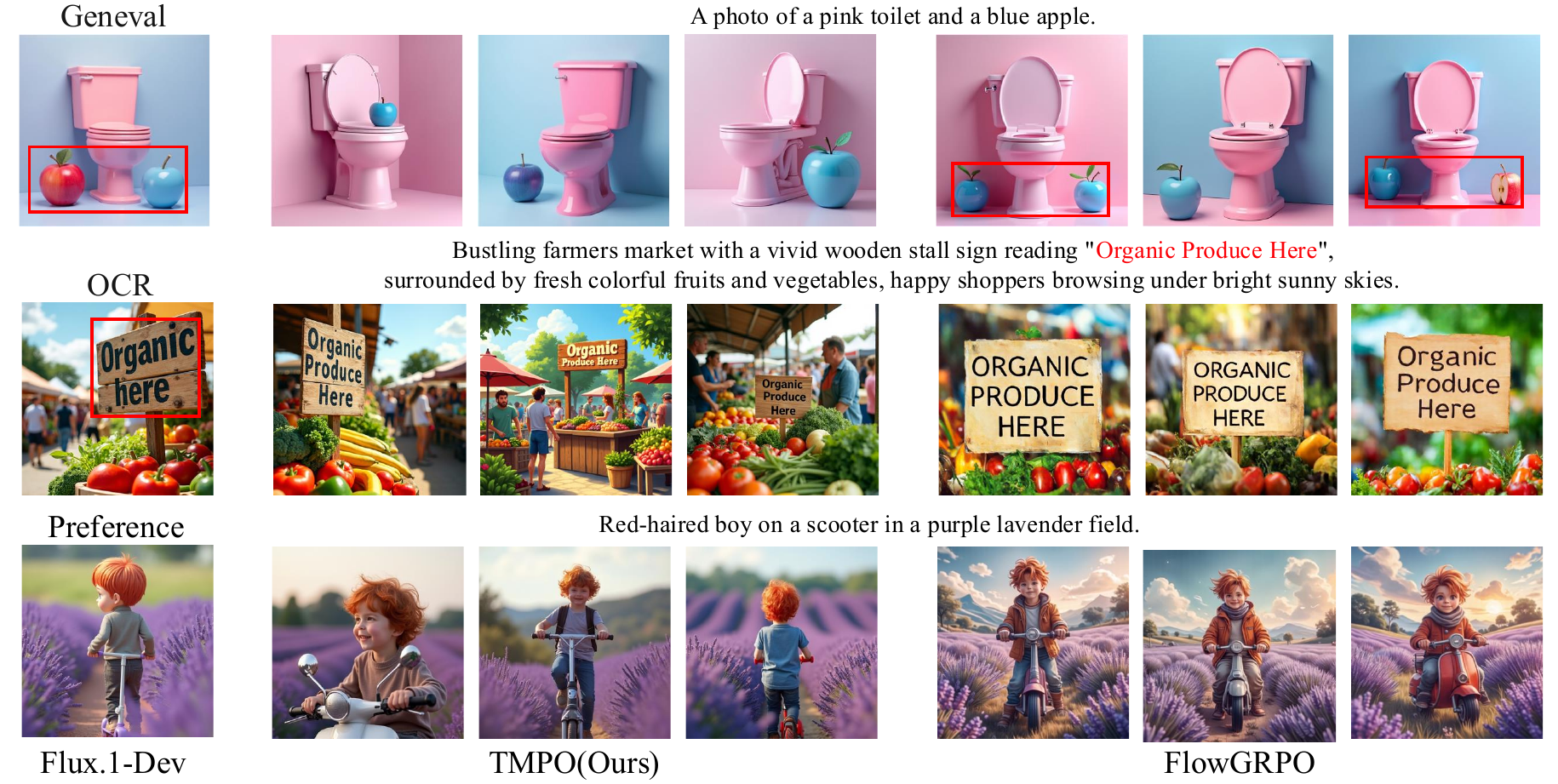}
  \caption{\textbf{Qualitative results of different alignment methods.} TMPO produces diverse, high-fidelity images that faithfully follow the text prompt.}
  \label{fig:demo}
\end{figure*}

\paragraph{Human preference alignment.}
Under PickScore only training, TMPO obtains the best PickScore and diversity, with competitive HPS-v2.1 and ImageReward. Under joint training (Table~\ref{tab:comparison_joint}), TMPO again achieves the best HPS-v2.1 and LGMD, confirming that distribution matching transfers better across unseen preference metrics. Figure~\ref{fig:training_curves} shows training curves across all three protocols; qualitative examples are shown in Figure~\ref{fig:demo}.

\begin{table}[h]
  \centering
  \caption{Performance on Compositional Image Generation, Visual Text Rendering, and Human Preference benchmarks on FLUX.1-dev, evaluated by task performance on task-specific test prompts and by preference scores and diversity on DrawBench prompts. Best values are \textbf{bold} and second-best are \underline{underlined}. \colorbox{gray!10}{Light gray rows} denote the pretrained baseline without RL training. \colorbox{gray!20}{Dark gray rows} denote our approach. \colorbox{proxyorange!18}{Orange cells} mark the proxy reward used for training in each section. ImgRwd: ImageReward; Cos.\,Div.: DINOv2-space cosine diversity.}
  \label{tab:main_results}
  \resizebox{\columnwidth}{!}{
    \begin{tabular}{lcccccccc}
      \toprule
      \multirow{2}{*}{\textbf{Method}} & \multirow{2}{*}{\textbf{Time (s)} $\downarrow$} & \multicolumn{2}{c}{\textbf{Task Metric} $\uparrow$} & \multicolumn{3}{c}{\textbf{Human Preference} $\uparrow$} & \multicolumn{2}{c}{\textbf{Diversity} $\uparrow$} \\
      \cmidrule(lr){3-4} \cmidrule(lr){5-7} \cmidrule(lr){8-9}
      & & \textbf{GenEval} & \textbf{OCR Acc.} & \textbf{HPS-v2.1} & \textbf{ImgRwd} & \textbf{PickScore} & \textbf{LGMD} & \textbf{Cos.\,Div.} \\
      \midrule
      \multicolumn{9}{c}{\textit{Compositional Image Generation}} \\
      \midrule
      \rowcolor{gray!10} FLUX.1-dev & --- & \cellcolor{proxyorange!18} 0.647 & --- & 0.301 & \underline{1.099} & \underline{22.301} & -0.031 & 0.211 \\
      Flow-GRPO & 160.7 & \cellcolor{proxyorange!18} \underline{0.943} & --- & 0.285 & 1.078 & 22.097 & -0.092 & 0.201 \\
      MixGRPO & 150.8 & \cellcolor{proxyorange!18} 0.932 & --- & \textbf{0.310} & 1.086 & 21.850 & -0.271 & 0.183 \\
      TreeGRPO & \underline{125.9} & \cellcolor{proxyorange!18} 0.939 & --- & 0.278 & 1.079 & 21.540 & -0.285 & 0.181 \\
      GARDO & 165.3 & \cellcolor{proxyorange!18} 0.923 & --- & 0.289 & 1.091 & 22.245 & \underline{0.012} & \underline{0.238} \\
      \rowcolor{gray!20}
      \textbf{TMPO (Ours)} & \textbf{91.9} & \cellcolor{proxyorange!18} \textbf{0.949} & --- & \underline{0.306} & \textbf{1.159} & \textbf{22.901} & \textbf{0.131} & \textbf{0.247} \\
      \midrule
      \multicolumn{9}{c}{\textit{Visual Text Rendering}} \\
      \midrule
      \rowcolor{gray!10} FLUX.1-dev & --- & --- & \cellcolor{proxyorange!18} 0.591 & \underline{0.292} & \textbf{1.121} & 21.968 & -0.040 & 0.215 \\
      Flow-GRPO & 121.3 & --- & \cellcolor{proxyorange!18} \textbf{0.940} & 0.291 & 1.100 & 21.365 & -0.092 & 0.208 \\
      MixGRPO & 116.2 & --- & \cellcolor{proxyorange!18} 0.919 & 0.282 & 1.098 & 20.878 & -0.302 & 0.179 \\
      TreeGRPO & \underline{93.4} & --- & \cellcolor{proxyorange!18} 0.921 & 0.290 & 1.110 & 21.099 & -0.292 & 0.180 \\
      GARDO & 123.8 & --- & \cellcolor{proxyorange!18} 0.931 & 0.291 & 1.085 & \underline{22.087} & \underline{0.064} & \underline{0.228} \\
      \rowcolor{gray!20}
      \textbf{TMPO (Ours)} & \textbf{76.3} & --- & \cellcolor{proxyorange!18} \underline{0.935} & \textbf{0.310} & \underline{1.111} & \textbf{22.309} & \textbf{0.110} & \textbf{0.241} \\
      \midrule
      \multicolumn{9}{c}{\textit{Human Preference Alignment}} \\
      \midrule
      \rowcolor{gray!10} FLUX.1-dev & --- & --- & --- & 0.310 & 1.119 & \cellcolor{proxyorange!18} 22.604 & \underline{-0.056} & \underline{0.214} \\
      Flow-GRPO & 108.9 & --- & --- & \textbf{0.377} & 1.600 & \cellcolor{proxyorange!18} \underline{24.210} & -0.107 & 0.209 \\
      MixGRPO & 103.5 & --- & --- & 0.367 & \textbf{1.621} & \cellcolor{proxyorange!18} 23.934 & -0.232 & 0.183 \\
      TreeGRPO & \underline{79.1} & --- & --- & 0.369 & 1.580 & \cellcolor{proxyorange!18} 23.657 & -0.284 & 0.176 \\
      GARDO & 112.7 & --- & --- & 0.343 & 1.571 & \cellcolor{proxyorange!18} 23.959 & -0.060 & 0.212 \\
      \rowcolor{gray!20}
      \textbf{TMPO (Ours)} & \textbf{68.3} & --- & --- & \underline{0.373} & \underline{1.610} & \cellcolor{proxyorange!18} \textbf{24.277} & \textbf{0.204} & \textbf{0.252} \\
      \bottomrule
    \end{tabular}
  }
\end{table}

\paragraph{Efficiency and diversity.}
TMPO improves the Pareto frontier between alignment, diversity, and training cost. Because trajectory log-probabilities accumulate only over $T{=}3$ stochastic branch points, gradient back-propagation is confined to these transitions rather than the full denoising horizon; combined with prefix sharing that amortizes forward computation, TMPO reduces iteration time by $\sim$20\% relative to TreeGRPO across three single-reward settings (Table~\ref{tab:main_results}; joint training in Table~\ref{tab:comparison_joint}). More importantly, the LGMD gap is consistent across all settings: GRPO-style baselines often raise reward scores while driving LGMD negative, whereas TMPO keeps LGMD positive in every setting.

\subsection{Ablation study}

We ablate major components of TMPO under the PickScore only protocol (Table~\ref{tab:ablation}), covering three categories: \textit{objective} (reward temperature $\beta$), \textit{balance granularity} (trajectory vs.\ step-wise), and \textit{sampling strategy} (tree structure, dynamic branching).

\begin{table}[hbt]
  \caption{Ablation study on FLUX.1-dev under the PickScore only protocol. Time denotes iteration time in seconds.}
  \label{tab:ablation}
  \centering
  \resizebox{\linewidth}{!}{%
  \begin{tabular}{lcccccc}
      \toprule
      \textbf{Variant} & \textbf{Time} $\downarrow$ & \textbf{HPS-v2.1} $\uparrow$ & \textbf{ImageReward} $\uparrow$ & \textbf{PickScore} $\uparrow$ & \textbf{LGMD} $\uparrow$ & \textbf{Cos.\,Div.} $\uparrow$ \\
      \midrule
      \textbf{TMPO (Full)} & \textbf{68.3} & \textbf{0.373} & \textbf{1.610} & \textbf{24.277} & \textbf{0.204} & \textbf{0.252} \\
      \midrule
      w/o Reward Temp.\ $\beta$ ($\beta{=}1$) & \underline{68.5} & 0.352 & 1.501 & 23.891 & \underline{0.199} & \underline{0.250} \\
      Step-wise Balance (Detailed) & 87.2 & 0.339 & 1.139 & 23.107 & 0.012 & 0.223 \\
      w/o Tree Structure ($K{=}27$) & 123.7 & \underline{0.369} & \underline{1.598} & \underline{24.109} & 0.173 & 0.245 \\
      w/o Dynamic Branching & 78.7 & 0.361 & 1.553 & 23.878 & 0.121 & 0.239 \\
      \bottomrule
    \end{tabular}%
  }
\end{table}

The results confirm the contribution of each component.
\textbf{w/o Reward Temp.\ $\beta$} removes the $\beta$-warmup schedule and fixes $\beta{=}1$ throughout training. Without progressive sharpening, the target distribution remains flat, slowing reward improvement (PickScore $24.28 \to 23.89$); however, the mild optimization pressure preserves diversity well (LGMD $0.199$, Cos.\,Div.\ $0.250$).
\textbf{Step-wise detailed balance} enforces per-step flow matching, causing severe loss oscillation and convergence difficulty. The substantial drop in ImageReward ($1.61 \to 1.14$) and near-zero LGMD ($0.012$) show that local supervision fragments credit assignment across the trajectory.
\textbf{w/o Tree Structure} uses $K{=}27$ independent rollouts and achieves comparable alignment, confirming the generality of Softmax-TB; however, iteration time increases by $81\%$ due to the loss of prefix sharing.
\textbf{w/o Dynamic Branching} fixes branch positions throughout training, resulting in lower diversity ($\Delta\text{LGMD} = -0.083$, $\Delta\text{Cos.\,Div.} = -0.013$) and higher iteration time ($78.7$\,s vs.~$68.3$\,s).

\section{Conclusion}
We presented TMPO, which reformulates diffusion RL alignment as trajectory-level distribution matching via Softmax Trajectory Balance, a partition-free objective whose advantage characterizes the forward KL divergence, inheriting mode-covering diversity guarantees without auxiliary regularization. Combined with prefix-sharing tree rollouts and dynamic branching, TMPO achieves the best reward--diversity trade-off across all evaluated settings while reducing iteration time by up to 27\%.

\paragraph{Limitations \& Future Work.}
This work focuses on T2I alignment; two natural extensions are:
(1)~the Boltzmann target quality is upper-bounded by the underlying reward model; pairing TMPO with stronger reward models may yield further improvements, and
(2)~extending trajectory-level distribution matching to video generation, 3D synthesis, and robotic action generation.

\clearpage
\beginappendix

\startcontents[app]
\printcontents[app]{l}{1}{%
}{}

\vspace{1em}

\noindent Our Appendix consists of 5 sections. Readers can click on each section number to navigate to the corresponding section:
\vspace{-0.6em}
\begin{itemize}[leftmargin=2.0em]
    \item Section~\ref{appendix:theoretical_derivations} and Section~\ref{appendix:gradient_analysis} provide mathematical derivations of Softmax-TB and importance sampling analysis.
    \item Section~\ref{appendix:tree_sampling} details the dynamic stochastic tree sampling algorithm.
    \item Section~\ref{app:sec:exp-setup-details} presents details about our experimental setup.
    \item Section~\ref{app:sec:extended-exp-results} provides additional experimental results, including joint preference training, generalization to SD3.5-Medium, qualitative comparisons with baselines, and evolution of evaluation images across training steps.
\end{itemize}
\vspace{-0.4em}

\clearpage
\section{Theoretical Derivations of Softmax-TB}
\label{appendix:theoretical_derivations}

\subsection{Boltzmann Target and Partition-Free Softmax Matching}

This subsection provides detailed proofs of the two foundational results used in the main text: (i) the Boltzmann distribution as the unique max-entropy target, and (ii) the elimination of the partition function via within-group normalization.

\subsubsection{Max-Entropy Derivation of the Boltzmann Distribution}

In the TMPO framework, we transform the raw reward $R$ into an exponential form $R \mapsto \exp(\beta R)$. The theoretical justification for this mapping is that it represents the unique solution that maximizes the distributional entropy (exploration diversity) subject to a fixed expected reward constraint.

\begin{proof}
Consider the search for an optimal probability distribution $p^*(\tau)$ in trajectory space. Our objective is to maximize the Shannon entropy $H(p)$ while satisfying the mean reward constraint $\mathbb{E}_{p}[R(\tau)] = \bar{R}$:
\begin{equation}
\begin{aligned}
    \max_{p} \quad & H(p) = -\sum_{\tau} p(\tau) \log p(\tau) \\
    \text{s.t.} \quad & \sum_{\tau} p(\tau) R(\tau) = \bar{R}, \quad \sum_{\tau} p(\tau) = 1,
\end{aligned}
\end{equation}
where $\bar{R}$ denotes a predefined expected reward level. Introducing the Lagrangian multipliers $\beta$ (inverse temperature) and $\lambda$, we construct the functional:
\begin{equation}
    \mathcal{L}(p, \beta, \lambda) = -\sum_{\tau} p(\tau) \log p(\tau) + \beta \left( \sum_{\tau} p(\tau) R(\tau) - \bar{R} \right) + \lambda \left( \sum_{\tau} p(\tau) - 1 \right).
\end{equation}
Setting the functional derivative with respect to $p(\tau)$ to zero:
\begin{equation}
    \frac{\delta \mathcal{L}}{\delta p(\tau)} = -\log p(\tau) - 1 + \beta R(\tau) + \lambda = 0.
\end{equation}
Solving for $p^*(\tau)$ yields $p^*(\tau) = \exp(\lambda - 1) \exp(\beta R(\tau))$. By enforcing the normalization constraint $\sum_{\tau} p(\tau) = 1$, we define the partition function $Z_\beta = \sum_{\tau'} \exp(\beta R(\tau'))$, resulting in:
\begin{equation}
    p^*(\tau) = \frac{1}{Z_\beta} \exp(\beta R(\tau)).
\end{equation}
This derivation proves that the Boltzmann distribution is the unique maximum-entropy distribution consistent with a given expected reward level.
\end{proof}

\subsubsection{Elimination of the Partition Function in Tree Topologies}

In standard GFlowNets~\cite{bengio2021flow}, the Trajectory Balance (TB) objective~\cite{malkin2022trajectory} is defined as $Z \prod_{t} P_F = R \prod_{t} P_B$. In the tree-structured sampling of TMPO, each state possesses a unique parent, which implies that the backward path probability is strictly $P_B(\tau) = 1$. Consequently, the flow matching condition simplifies to $P_\theta(\tau) = R(\tau) / Z_\theta$.

Consider a group of $K$ trajectories sampled from the same branching point $x_{\text{split}}$. Since these trajectories share a common prefix path, the global partition function $Z_\theta$ is constant across the group. The relative probability of trajectory $\tau_i$ within the group is given by:
\begin{equation}
    \frac{P_{\theta}(\tau_i)}{\sum_{j=1}^K P_{\theta}(\tau_j)} = \frac{R(\tau_i) / Z_{\theta}}{\sum_{j=1}^K R(\tau_j) / Z_{\theta}} = \frac{R(\tau_i)}{\sum_{j=1}^K R(\tau_j)}.
\end{equation}
Substituting the exponential reward mapping $R \to \exp(\beta R)$ yields the softmax matching format:
\begin{equation}
    \frac{P_{\theta}(\tau_i)}{\sum_{j=1}^K P_{\theta}(\tau_j)} = \frac{\exp(\beta R_i)}{\sum_{j=1}^K \exp(\beta R_j)}.
\end{equation}

\subsection{Forward KL Characterization of Softmax-TB (Extended Proof)}
\label{appendix:forward_kl}

We now formally characterize the mode-covering property of the Softmax-TB advantage and its connection to forward KL divergence, providing complete proofs and quantitative guarantees that supplement the mode-covering equilibrium result in Section~\ref{prop:mode_covering}.

\subsubsection{Formal Statement and Proof}

Define the within-group Boltzmann target $q_i \triangleq \exp(\beta R_i)/\sum_{j=1}^K \exp(\beta R_j)$ and the within-group policy distribution $p_i \triangleq P_\theta(\tau_i)/\sum_{j=1}^K P_\theta(\tau_j)$. Both $q$ and $p$ are valid probability distributions over the $K$ trajectories, i.e., $q_i \geq 0$, $p_i > 0$, $\sum_i q_i = \sum_i p_i = 1$.

The Softmax-TB advantage is $A_i = \log q_i - \log p_i$. We now prove:

\begin{equation}
\sum_{i=1}^K q_i \, A_i = \sum_{i=1}^K q_i (\log q_i - \log p_i) = D_{\mathrm{KL}}(q \| p) \geq 0.
\end{equation}

\textbf{Step 1 (Substitution).} By definition of the KL divergence:
\begin{equation}
D_{\mathrm{KL}}(q \| p) = \sum_{i=1}^K q_i \log \frac{q_i}{p_i} = \sum_{i=1}^K q_i (\log q_i - \log p_i) = \sum_{i=1}^K q_i A_i.
\end{equation}

\textbf{Step 2 (Non-negativity via Gibbs' inequality).} By Jensen's inequality applied to the concave function $\log(\cdot)$:
\begin{equation}
\sum_{i=1}^K q_i \log \frac{p_i}{q_i} \leq \log \sum_{i=1}^K q_i \cdot \frac{p_i}{q_i} = \log \sum_{i=1}^K p_i = \log 1 = 0.
\end{equation}
Since $D_{\mathrm{KL}}(q \| p) = -\sum_i q_i \log(p_i / q_i)$, it follows that $D_{\mathrm{KL}}(q \| p) \geq 0$.

\textbf{Step 3 (Equality condition).} Equality holds in Jensen's inequality if and only if $p_i / q_i$ is constant for all $i$ with $q_i > 0$. Combined with the normalization constraints $\sum_i p_i = \sum_i q_i = 1$, this constant must be $1$, i.e., $p_i = q_i$ for all $i$.

\textbf{Step 4 (Partition function independence).} The within-group normalization is critical: $q_i = \exp(\beta R_i) / \sum_j \exp(\beta R_j)$ and $p_i = P_\theta(\tau_i) / \sum_j P_\theta(\tau_j)$. The global partition function $Z_\beta = \sum_{\text{all } \tau} \exp(\beta R(\tau))$ and the global policy normalization $Z_\theta$ both cancel in the ratio. The forward KL is computed \emph{exactly} over the $K$ observed trajectories without requiring access to the full trajectory space.

\subsubsection{Gradient Direction Under Detached Advantage}

When the advantage is detached ($A_i^{\bot}$), the policy gradient of the Softmax-TB loss takes the form:
\begin{equation}\label{eq:softmax_tb_gradient}
\nabla_\theta \mathcal{L} = -\frac{1}{K} \sum_{i=1}^K A_i^{\bot} \cdot \nabla_\theta \log P_\theta(\tau_i).
\end{equation}
We show that this gradient direction is sign-consistent with the gradient of the forward KL $D_{\mathrm{KL}}(q \| p_\theta)$.

\textbf{Exact forward KL gradient.} Let $p_i(\theta) = P_\theta(\tau_i) / \sum_j P_\theta(\tau_j)$. By the chain rule:
\begin{equation}\label{eq:exact_fkl_grad}
\nabla_\theta D_{\mathrm{KL}}(q \| p_\theta) = -\sum_{i=1}^K q_i \, \nabla_\theta \log p_i(\theta).
\end{equation}
Expanding $\nabla_\theta \log p_i = \nabla_\theta \log P_\theta(\tau_i) - \sum_j p_j \nabla_\theta \log P_\theta(\tau_j)$, we obtain:
\begin{equation}
\nabla_\theta D_{\mathrm{KL}}(q \| p_\theta) = -\sum_{i=1}^K (q_i - p_i) \, \nabla_\theta \log P_\theta(\tau_i),
\end{equation}
where the baseline $\sum_j p_j \nabla_\theta \log P_\theta(\tau_j)$ has been subtracted. Comparing the two gradients, TMPO uses $A_i/K = \log(q_i/p_i)/K$ as the per-trajectory weight, while the exact forward KL uses $(q_i - p_i)$. Although the magnitudes differ, the signs are strictly identical: $\mathrm{sign}(A_i) = \mathrm{sign}(\log(q_i/p_i)) = \mathrm{sign}(q_i - p_i)$. This means:
\begin{itemize}
\item $A_i > 0$ ($q_i > p_i$): trajectory $\tau_i$ is under-covered $\Rightarrow$ both gradients increase $P_\theta(\tau_i)$.
\item $A_i < 0$ ($q_i < p_i$): trajectory $\tau_i$ is over-represented $\Rightarrow$ both gradients decrease $P_\theta(\tau_i)$.
\end{itemize}
The detached gradient omits the baseline term, introducing magnitude bias but preserving the sign of each trajectory's gradient contribution. This is analogous to REINFORCE without baseline: the per-trajectory gradient direction remains correct, while the baseline would reduce variance at the cost of cross-trajectory coupling. In finite groups, the omitted baseline $\sum_j p_j \nabla_\theta \log P_\theta(\tau_j)$ does not vanish exactly, so the estimator is biased in magnitude but sign-consistent with the exact forward KL gradient.

\textbf{Role of the IS ratio.} In the full TMPO loss, the gradient flows through the bias-corrected IS ratio $\hat{\rho}_i = P_\theta(\tau_i)/P_{\mathrm{old}}(\tau_i)$ rather than directly through $\log P_\theta(\tau_i)$. The IS ratio serves as the gradient carrier for off-policy correction; it is \emph{not} a substitute for the Boltzmann weight $q_i$ that appears in the exact forward KL gradient. These two quantities are functionally independent: $\hat{\rho}_i$ corrects for the distribution shift between successive policy updates, while $A_i^{\bot}$ provides the directional signal for distribution matching.

\subsubsection{Mode-Covering vs.\ Mode-Seeking: Formal Characterization}

The forward and reverse KL divergences impose fundamentally different penalties on the mismatch between $q$ and $p$. We formalize this distinction below.

\textbf{Forward KL} $D_{\mathrm{KL}}(q \| p) = \sum_i q_i \log(q_i / p_i)$: For any trajectory $i$ with $q_i > 0$, the penalty $q_i \log(q_i / p_i) \to +\infty$ as $p_i \to 0$. This infinite cost forces $p$ to assign non-zero probability mass everywhere that $q$ is positive, producing \emph{mode-covering} behavior. Formally, the minimizer $p^* = \arg\min_p D_{\mathrm{KL}}(q \| p)$ satisfies $\mathrm{supp}(p^*) \supseteq \mathrm{supp}(q)$.

\textbf{Reverse KL} $D_{\mathrm{KL}}(p \| q) = \sum_i p_i \log(p_i / q_i)$: When $p_i = 0$, the contribution is exactly $0$ regardless of $q_i$. The policy can collapse to a single mode without penalty, producing \emph{mode-seeking} behavior. The minimizer concentrates on high-$q$ regions: $p^* = \arg\min_p D_{\mathrm{KL}}(p \| q)$ typically yields $\mathrm{supp}(p^*) \subseteq \mathrm{supp}(q)$.

\textbf{Quantitative bound via Pinsker's inequality.} The mode-covering guarantee can be strengthened by Pinsker's inequality, which relates the KL divergence to the total variation distance:
\begin{equation}
\mathrm{TV}(q, p) \triangleq \frac{1}{2}\sum_{i=1}^K |q_i - p_i| \leq \sqrt{\frac{1}{2} D_{\mathrm{KL}}(q \| p)}.
\end{equation}
Therefore, when Softmax-TB drives $D_{\mathrm{KL}}(q \| p) \to 0$, the policy distribution $p$ converges to $q$ in total variation, guaranteeing that no mode is under-covered by more than $\sqrt{D_{\mathrm{KL}}(q\|p)/2}$ in absolute probability.

\textbf{Implication for Softmax-TB.} Since the Boltzmann target $q$ assigns strictly positive probability to all $K$ trajectories ($\exp(\beta R_i) > 0$ for any finite $R_i$), minimizing $D_{\mathrm{KL}}(q \| p)$ requires $p_i > 0$ for all $i$. Moreover, by Pinsker's inequality, bounding the forward KL below $\delta$ ensures $|p_i - q_i| \leq \sqrt{\delta/2}$ for every trajectory. This provides a \emph{quantitative} diversity guarantee that standard GRPO-style reward maximization (reverse KL) cannot offer.

\subsubsection{Monotonic Descent Under Exact Updates}

The Softmax-TB loss and the within-group forward KL share the same unique fixed point $p_i = q_i$ for all $i$. Under idealized exact updates, we establish that the forward KL is monotonically non-increasing.

\textbf{Claim.} Let $\theta^{(n)}$ denote the parameters after iteration $n$. If the update rule is $\theta^{(n+1)} = \arg\min_\theta D_{\mathrm{KL}}(q^{(n)} \| p_\theta)$ where $q^{(n)}$ is the Boltzmann target computed from the group sampled at iteration $n$, then:
\begin{equation}
D_{\mathrm{KL}}(q^{(n)} \| p_{\theta^{(n+1)}}) \leq D_{\mathrm{KL}}(q^{(n)} \| p_{\theta^{(n)}}).
\end{equation}

\begin{proof}
By definition, $\theta^{(n+1)}$ minimizes $D_{\mathrm{KL}}(q^{(n)} \| p_\theta)$ over $\theta$. Since $\theta^{(n)}$ is a feasible point, we have $D_{\mathrm{KL}}(q^{(n)} \| p_{\theta^{(n+1)}}) \leq D_{\mathrm{KL}}(q^{(n)} \| p_{\theta^{(n)}})$. We note that this single-step guarantee is tautological under the exact-minimization assumption; it serves to establish the fixed-point structure, not to claim practical convergence rates. In practice, TMPO performs gradient steps rather than exact minimization, and the trust-region clipping ensures that each step remains close to $\theta^{(n)}$, bounding the gap between the approximate and exact updates.
\end{proof}

\subsubsection{Connection to Within-Group Distribution Matching}

A natural concern is that Softmax-TB samples from $p_{\text{old}}$ rather than $q_\beta$, seemingly preventing the computation of the log-ratio advantage $A_i = \log(q_i/p_i)$. We clarify that this is a question of \emph{computational feasibility}, not of the loss identity.

In each training iteration, TMPO generates a fixed group of $K$ trajectories $\{\tau_1, \ldots, \tau_K\}$ from the current tree sampler. Within this group, both $q$ and $p$ are discrete distributions over the \emph{same} $K$ atoms. The advantage $A_i = \log(q_i/p_i)$ is computed exactly over these $K$ fully observed atoms; no sampling from $q$ is required. This is analogous to supervised cross-entropy loss: given a fixed dataset of $K$ labeled examples, the cross-entropy $H(q, p) = -\sum_i q_i \log p_i$ can be evaluated without sampling from the label distribution $q$, because all $K$ labels are observed.

Formally, define the within-group forward KL at iteration $n$ as:
\begin{equation}
\mathcal{F}_n(\theta) \triangleq D_{\mathrm{KL}}(q^{(n)} \| p_\theta^{(n)}) = \sum_{i=1}^K q_i^{(n)} \log \frac{q_i^{(n)}}{p_i(\theta)},
\end{equation}
where $q^{(n)}$ and $p^{(n)}$ are both normalized over the $K$ trajectories sampled at iteration $n$. Note that any group-based method (including GRPO) could in principle compute this quantity; the distinction is that TMPO's advantage $A_i = \log(q_i/p_i)$ \emph{uses} this log-ratio as its optimization signal, whereas reward-maximization methods discard the policy probability $p_i$ entirely. Within-group normalization converts the intractable global distribution matching problem into a tractable local one, ensuring that the distribution-aware advantage can be computed exactly without access to the global Boltzmann distribution $q_\beta$. Each group covers a finite subset of the trajectory space; stochastic tree exploration (SDE noise injection and the dynamic Beta branching schedule; Section~\ref{appendix:tree_sampling}) diversifies successive groups so that the cumulative effect of exact within-group matching progressively extends across the trajectory manifold.

\subsection{Advantage Structure Comparison with Standard Reward Maximization}
\label{appendix:reverse_kl_comparison}

Standard policy gradient methods, including TMPO, sample trajectories from the current policy $\pi_\theta$, placing them in a reverse KL sampling regime. The distinction lies not in the sampling distribution but in the advantage structure. We formalize the connection between reward maximization and reverse KL, then contrast the resulting advantage with the distribution-aware advantage of Softmax-TB.

\subsubsection{Reward Maximization as Reverse KL Minimization}

Consider the standard RL objective of maximizing expected reward:
\begin{equation}
\max_\theta \; \mathbb{E}_{\tau \sim \pi_\theta}[R(\tau)].
\end{equation}
Let $q(\tau) = \exp(R(\tau))/Z$ denote the Boltzmann target, so that $R(\tau) = \log q(\tau) + \log Z$. Substituting:
\begin{equation}
\mathbb{E}_{\pi_\theta}[R(\tau)] = \mathbb{E}_{\pi_\theta}[\log q(\tau)] + \log Z = -H(\pi_\theta, q) + \log Z,
\end{equation}
where $H(\pi_\theta, q) = -\sum_\tau \pi_\theta(\tau) \log q(\tau)$ is the cross-entropy. Using $D_{\mathrm{KL}}(\pi_\theta \| q) = H(\pi_\theta, q) - H(\pi_\theta)$:
\begin{equation}\label{eq:reward_reverse_kl}
\boxed{\mathbb{E}_{\pi_\theta}[R(\tau)] = -D_{\mathrm{KL}}(\pi_\theta \| q) - H(\pi_\theta) + \log Z.}
\end{equation}
Since $\log Z$ is a constant, maximizing expected reward is equivalent to minimizing $D_{\mathrm{KL}}(\pi_\theta \| q) + H(\pi_\theta)$. This reveals a compounding mode-collapse pressure: reward maximization not only minimizes the reverse KL (which is mode-seeking) but also minimizes the policy entropy, further concentrating probability mass. Ignoring the entropy contribution:
$\max_\theta \mathbb{E}[R] \equiv \min_\theta D_{\mathrm{KL}}(\pi_\theta \| q)$.

\subsubsection{Advantage Structure Comparison}

The fundamental distinction between TMPO and standard reward maximization lies in the advantage structure:

\textbf{Reward-maximization advantage} (z-scored reward, e.g.\ Flow-GRPO~\cite{liu2025flowgrpo}):
\begin{equation}
A_i^{\mathrm{GRPO}} = \frac{R_i - \mu_R}{\sigma_R},
\end{equation}
where $\mu_R$ and $\sigma_R$ are the within-group mean and standard deviation of the rewards. This is a \emph{linear function of $R_i$} that is entirely agnostic to the policy probability $p_i = P_\theta(\tau_i) / \sum_j P_\theta(\tau_j)$.

\textbf{Softmax-TB advantage} (log-ratio):
\begin{equation}
A_i^{\mathrm{TMPO}} = \log q_i - \log p_i = \log \frac{q_i}{p_i}.
\end{equation}
This is \emph{distribution-aware}: it simultaneously encodes both the Boltzmann target weight $q_i$ and the current policy probability $p_i$.

\subsubsection{Asymptotic Behavior Under Mode Dropping}

Consider the scenario where the policy drops a mode, i.e., $p_i \to 0$ for some trajectory $i$ with $q_i > 0$:

\textbf{Reward maximization:} The advantage $A_i^{\mathrm{GRPO}} = (R_i - \mu_R)/\sigma_R$ depends only on $R_i$ and the group statistics. As $p_i \to 0$, the advantage does not change; the gradient signal for trajectory $i$ is unaffected by the policy's diminishing probability on it. Formally, the reverse KL contribution $p_i \log(p_i/q_i) \to 0$ as $p_i \to 0$, confirming that mode-dropping incurs zero penalty.

\textbf{TMPO:} The advantage $A_i^{\mathrm{TMPO}} = \log(q_i/p_i) \to +\infty$ as $p_i \to 0$. This divergent correction signal forces the policy to restore probability mass on the dropped mode. This behavior directly mirrors the forward KL: $q_i \log(q_i/p_i) \to +\infty$ as $p_i \to 0$. We summarize the structural comparison between the two advantage functions in Table~\ref{tab:advantage_comparison}.

\begin{table}[ht]
\centering
\caption{Structural comparison of advantage functions and their KL divergence properties.}
\label{tab:advantage_comparison}
\resizebox{\columnwidth}{!}{
\begin{tabular}{lcc}
\toprule
& \textbf{Reward Maximization} & \textbf{TMPO (Softmax-TB)} \\
\midrule
Advantage & $A_i \propto R_i - \bar{R}$ (linear in reward) & $A_i = \log(q_i/p_i)$ (log-ratio) \\
Aware of $p_i$? & No & Yes \\
$p_i \to 0$ behavior & $A_i$ unchanged & $A_i \to +\infty$ \\
KL regime & Reverse $D_{\mathrm{KL}}(p \| q)$ & Inherits forward $D_{\mathrm{KL}}(q \| p)$ asymptote \\
Equilibrium & $\pi_\theta \to \arg\max R$ & $p_i = q_i \;\forall i$ \\
Mode-covering guarantee & Requires external regularization & Built into advantage structure \\
\bottomrule
\end{tabular}
}
\end{table}

Since the reverse KL assigns zero penalty to dropped modes ($p_i\log(p_i/q_i) \to 0$), reward-maximization methods must rely on external mechanisms---KL penalties against a reference policy, or heuristic advantage reweighting---to counteract mode collapse. These mechanisms are either agnostic to the reward landscape (KL penalty) or unable to detect modes that have already been dropped (reweighting). The forward KL asymptote built into TMPO's advantage provides this guarantee structurally: any mode with $p_i < q_i$ receives a corrective signal proportional to $\log(q_i/p_i)$, without auxiliary losses or hyperparameters.

\section{Importance Sampling, RatioNorm, and Gradient Analysis}
\label{appendix:gradient_analysis}

This section provides full derivations of the importance sampling decomposition and bias correction used in the Softmax-TB objective (Equation~\eqref{eq:softmax_tb_loss}), and analyzes the resulting trust-region properties.

\subsection{Importance Sampling Ratio Decomposition}

To enable multiple policy updates per batch, TMPO requires importance sampling (IS) correction. In the tree-structured sampling, the path log-probability decomposes into $T$ per-step transition probabilities at the stochastic branch points: $\log P_\theta(\tau_i) = \sum_{t=1}^T \log \pi_\theta(x_{t-1}^{(i)} \mid x_t^{(i)})$. Consequently, the trajectory-level log IS ratio admits a per-step decomposition:
\begin{equation}
\log \rho_i = \log P_\theta(\tau_i) - \log P_{\text{old}}(\tau_i) = \sum_{t=1}^T \underbrace{\left[\log \pi_\theta(x_{t-1}\mid x_t) - \log \pi_{\text{old}}(x_{t-1}\mid x_t)\right]}_{\triangleq \, \log w_{i,t}}
\end{equation}

Under the Gaussian transition kernel $\pi_\theta(x_{t-1}\mid x_t) = \mathcal{N}(x_{t-1};\, \mu_\theta(x_t),\, \sigma_t^2 \Delta t \, \mathbf{I})$, each per-step log-ratio decomposes as:
\begin{equation}
\log w_{i,t} = \frac{\Delta \mu_\theta \cdot \epsilon}{\sigma_t \sqrt{\Delta t}} - \frac{\|\Delta \mu_\theta\|^2}{2\sigma_t^2 \Delta t}
\end{equation}
where $\Delta \mu_\theta \triangleq \mu_\theta(x_t) - \mu_{\text{old}}(x_t)$ is the mean shift and $\epsilon \sim \mathcal{N}(0, \mathbf{I})$ is the sampled noise. The second term introduces a deterministic negative shift: $\mathbb{E}[\log w_{i,t}] = -\|\Delta\mu_\theta\|^2 / (2\sigma_t^2 \Delta t) < 0$. Note that the IS ratio itself is unbiased in ratio space ($\mathbb{E}[w_{i,t}] = 1$ by construction), and the negative log-expectation is a natural consequence of Jensen's inequality ($\mathbb{E}[\log w] < \log \mathbb{E}[w] = 0$). However, this deterministic negative shift in log-space causes the clipping interval $[1{-}\varepsilon, 1{+}\varepsilon]$ to be asymmetrically effective, suppressing gradient signals from high-reward trajectories.

\textbf{Log-space centering (RatioNorm).} Following GRPO-Guard~\cite{wang2025grpo}, TMPO removes the deterministic negative shift to center the log-ratio distribution around zero:
\begin{equation}\label{eq:ratio}
\log \hat{w}_{i,t} = \log w_{i,t} + \frac{\|\Delta \mu_\theta\|^2}{2\sigma_t^2 \Delta t} = \frac{\Delta \mu_\theta \cdot \epsilon}{\sigma_t \sqrt{\Delta t}}
\end{equation}
The resulting centered log-ratio has zero mean ($\mathbb{E}[\log\hat{w}_{i,t}]=0$ since $\mathbb{E}[\epsilon]=0$), restoring symmetric clipping behavior. Note that this centering implies $\mathbb{E}[\hat{w}_{i,t}] > 1$ by Jensen's inequality; the operation trades unbiasedness in ratio space for symmetry in log-space, which is the relevant domain for the clipping operator. Unlike the full GRPO-Guard formulation which additionally rescales by $\sigma_t\sqrt{\Delta t}$ for variance normalization, TMPO retains the unscaled form to preserve gradient magnitude in the few-step ($T \leq 5$) SDE regime. During backpropagation, the bias correction term is treated as a detached constant while $\log w_{i,t}$ retains its gradient.

\textbf{Trajectory-level aggregation.} The trajectory-level log-ratio is constructed as $\log \hat{w}_i = \sum_{t=1}^T \log \hat{w}_{i,t}$. In TMPO's tree sampler, $T \leq 5$ stochastic steps are used, so the sum remains $O(1)$ in practice and does not saturate the clipping interval $[1{-}\varepsilon, 1{+}\varepsilon]$. This direct summation preserves the full per-step gradient signal, which is important in the few-step SDE regime where time-averaging ($1/T$) would attenuate the gradient contribution of each stochastic transition.

\subsection{Trajectory-Level Gradient Decomposition}

The complete Softmax-TB loss takes the form:
\begin{equation}
\mathcal{L}_{\text{TB}}(\theta) = -\frac{1}{K} \sum_{i=1}^K \min\!\left(\hat{w}_i(\theta)\, A_i^{\bot},\; \text{clip}\!\left(\hat{w}_i(\theta),\, 1{-}\varepsilon,\, 1{+}\varepsilon\right) A_i^{\bot}\right)
\end{equation}
where $\hat{w}_i(\theta) = \exp\!\left(\sum_{t=1}^T \log \hat{w}_{i,t}(\theta)\right)$ is the bias-corrected importance ratio that depends on $\theta$, and $A_i^{\bot}$ is the detached Softmax-TB advantage treated as a constant with respect to $\theta$. The $\min$ selects whichever term yields the lower (more pessimistic) objective. Since no gradient flows through $A_i^{\bot}$, the gradient of the active branch admits a single-term decomposition:
\begin{equation}
\nabla_\theta \mathcal{L}_{\text{TB}}(\theta) = -\frac{1}{K} \sum_{i=1}^K \nabla_\theta \hat{w}_{i}^{\text{active}}(\theta) \cdot A_i^{\bot}
\end{equation}
where $\hat{w}_i^{\text{active}}$ denotes whichever of $\hat{w}_i$ or $\text{clip}(\hat{w}_i)$ is selected by the $\min$ operator. The detached advantage $A_i^{\bot}$ provides the directional signal, while the importance ratio is the sole carrier of gradient information. Crucially, since the bias correction terms are constants with respect to $\theta$, we have $\nabla_\theta \log \hat{w}_i = \sum_t \nabla_\theta \log \pi_\theta(x_{t-1} \mid x_t)$, which depends only on the log-probability of trajectory $\tau_i$. This ensures that each trajectory's gradient contribution is independent of all other trajectories within the group.

\subsection{Gradient Behavior in Unclipped and Clipped Regions}

The trust-region property arises from how the clipping operator modulates the gradient based on the magnitude of the importance ratio.

\textbf{Case 1: Within trust region ($\hat{w}_i \in [1{-}\varepsilon, 1{+}\varepsilon]$).} \\
Both branches of the $\min$ coincide and the gradient becomes:
\begin{equation}
\nabla_\theta \mathcal{L}_{\text{TB}}(\theta) \Big|_{\text{unclipped}} = -\frac{1}{K} \sum_{i=1}^K \hat{w}_i(\theta) \cdot \nabla_\theta \log \hat{w}_i(\theta) \cdot A_i^{\bot}
\end{equation}
In this region, the off-policy update proceeds normally. A positive advantage $A_i^{\bot} > 0$ drives the gradient to increase the path probability of trajectory $\tau_i$, while $A_i^{\bot} < 0$ decreases it.

\textbf{Case 2: Beneficial over-update ($A_i^{\bot} > 0,\, \hat{w}_i > 1{+}\varepsilon$ or $A_i^{\bot} < 0,\, \hat{w}_i < 1{-}\varepsilon$).} \\
The policy has moved in the direction indicated by the advantage beyond the trust region. In this case the clipped term yields a \emph{lower} objective value, so $\min$ selects it. Since the clipped term has zero gradient with respect to $\theta$, the trajectory contributes no update, preventing further over-optimization.

\textbf{Case 3: Harmful deviation ($A_i^{\bot} > 0,\, \hat{w}_i < 1{-}\varepsilon$ or $A_i^{\bot} < 0,\, \hat{w}_i > 1{+}\varepsilon$).} \\
The policy has drifted in the \emph{opposite} direction to the advantage. The unclipped term now yields the lower objective, so $\min$ selects it and the gradient flows through the unclipped ratio $\hat{w}_i$, providing a corrective signal that pushes the policy back. This asymmetric behavior—blocking beneficial over-updates while allowing corrective ones—is the key difference from a naive $\text{clip}(\hat{w}_i) \cdot A_i^{\bot}$ formulation and is essential for robust trust-region control.

\subsection{The Necessity of RatioNorm for Symmetric Clipping}

Applying the clipping bounds $[1{-}\varepsilon, 1{+}\varepsilon]$ directly to the raw importance ratio $\rho_i = P_\theta(\tau_i) / P_{\text{old}}(\tau_i)$ is ineffective due to the systematic bias inherent in the per-step log-ratios. As derived in the IS Ratio Decomposition (Section~\ref{appendix:gradient_analysis}), the raw log-ratio contains a negative bias:
\begin{equation}
\mathbb{E}[\log w_{i,t}] = -\frac{\|\Delta \mu_\theta\|^2}{2\sigma_t^2 \Delta t} < 0
\end{equation}
This causes $\mathbb{E}[\log \rho_i] < 0$, so the distribution of $\rho_i$ concentrates below $1$: the majority of trajectories fall below the lower clipping bound $1 - \varepsilon$, while the upper bound $1 + \varepsilon$ is rarely reached. The resulting asymmetry undermines the trust-region mechanism: the clipping cannot prevent excessively large updates in the direction of decreasing trajectory probability.

Bias correction restores symmetry by removing the negative expectation from each per-step log-ratio:
\begin{equation}
\log \hat{w}_{i,t} = \log w_{i,t} + \frac{\|\Delta \mu_\theta\|^2}{2\sigma_t^2 \Delta t} = \frac{\Delta \mu_\theta \cdot \epsilon}{\sigma_t\sqrt{\Delta t}}
\end{equation}
The bias correction ensures $\mathbb{E}[\log \hat{w}_{i,t}] = 0$, centering the trajectory-level ratio $\hat{w}_i = \exp(\sum_t \log \hat{w}_{i,t})$ around $1$ and enabling the symmetric clipping interval $[1{-}\varepsilon, 1{+}\varepsilon]$ to function as intended.

\section{Dynamic Stochastic Tree Sampling}
\label{appendix:tree_sampling}

This section provides detailed derivations for the stochastic tree sampler described in Section~\ref{sec:tree_sampling}, including the SDE noise injection, the curriculum-guided Beta branching schedule, and the trajectory log-probability computation.

\subsection{SDE Noise Injection at Branch Points}

The denoising process in flow-matching models follows the ODE $\mathrm{d}x = v_\theta(x, t)\,\mathrm{d}t$, where $v_\theta$ is the learned velocity field and $t$ decreases from $1$ (pure noise) to $0$ (clean image). To introduce stochasticity at branch points, TMPO converts this ODE into an equivalent SDE by injecting Gaussian noise. At branch point $s_i$ with noise schedule value $\sigma_{s_i}$, the SDE transition is:
\begin{equation}
x_{s_i^-} = \mu_\theta(x_{s_i}, s_i) + \gamma_i \, \varepsilon, \quad \varepsilon \sim \mathcal{N}(0, \mathbf{I}),
\end{equation}
where $\mu_\theta$ is the deterministic drift (incorporating both the velocity field and SDE drift correction), and $\gamma_i$ is the noise magnitude:
\begin{equation}\label{eq:noise_magnitude}
\gamma_i = \eta_i \sqrt{\frac{\sigma_{s_i}}{1 - \sigma_{s_i}}} \cdot \sqrt{-\Delta t_{s_i}}.
\end{equation}
Here $\eta_i \in (0, 1]$ is a per-layer noise coefficient, $\sigma_{s_i}/(1 - \sigma_{s_i})$ is the inverse signal-to-noise ratio at timestep $s_i$, and $\Delta t_{s_i} = \sigma_{s_i^-} - \sigma_{s_i} < 0$ is the (negative) time step. The $\sqrt{\text{SNR}^{-1}}$ factor ensures that noise magnitude scales naturally with the diffusion schedule: more noise is injected at early high-noise steps, while late low-noise steps receive proportionally less perturbation.

The SDE drift $\mu_\theta$ includes a correction term to preserve the marginal distribution:
\begin{equation}
\mu_\theta = x_{s_i} \left(1 + \frac{\gamma_i^2}{2\sigma_{s_i}} \Delta t\right) + v_\theta(x_{s_i}, s_i) \left(1 + \frac{\gamma_i^2 (1 - \sigma_{s_i})}{2\sigma_{s_i}}\right) \Delta t,
\end{equation}
where the additional $\gamma_i^2$ terms arise from the Itô-to-Stratonovich conversion and ensure that the SDE marginals match those of the original ODE.

At each branch point, $B$ independent noise realizations $\{\varepsilon_b\}_{b=1}^B$ are drawn, producing $B$ child branches from the same parent state. With $T$ branch points and branching factor $B$, the tree produces $K = B^T$ terminal trajectories per prompt (e.g., $B=3, T=3 \Rightarrow K=27$).

\subsection{Curriculum-Guided Beta Branching Schedule}
\label{appendix:beta_schedule}

Branch positions determine where along the denoising trajectory the tree bifurcates. Early branching (high $\sigma$) creates globally diverse structures, while late branching (low $\sigma$) produces fine-grained variations that share semantic structure. TMPO uses a curriculum scheduler that gradually shifts branch positions from early to late as training progresses.

\subsubsection{Deterministic Curriculum Trajectory}

For each branch point $i \in \{1, 2, 3\}$, the curriculum defines early positions $e_i$ and late positions $l_i$ (hyperparameters). Given normalized training progress $p = \min(u / U, 1)$ where $u$ is the current step and $U$ the total steps, the deterministic curriculum mean is:
\begin{equation}
\mu_i(p) = e_i + (l_i - e_i) \cdot p.
\end{equation}
This linear interpolation shifts branch positions from broad early exploration to fine-grained late refinement.

\subsubsection{Stochastic Perturbation via Beta Distribution}

To prevent the policy from overfitting to a fixed tree geometry, each branch position is stochastically perturbed around the curriculum mean. We normalize the curriculum mean to the interval $(0, 1)$:
\begin{equation}
\bar{\mu}_i(p) = \frac{\mu_i(p) - s_{\min}}{s_{\max} - s_{\min}}, \quad \bar{\mu}_i \in (0, 1),
\end{equation}
where $s_{\min}$ and $s_{\max}$ are the minimum and maximum allowed step indices. The branch position is then sampled from a Beta distribution:
\begin{equation}
\xi_i \sim \mathrm{Beta}\!\left(\bar{\mu}_i \kappa,\; (1 - \bar{\mu}_i) \kappa\right),
\end{equation}
where $\kappa > 0$ is the concentration parameter controlling the spread of the perturbation.

\textbf{Properties of this parameterization.} The Beta distribution $\mathrm{Beta}(\alpha, \beta)$ with $\alpha = \bar{\mu}\kappa$ and $\beta = (1-\bar{\mu})\kappa$ has:
\begin{align}
\mathbb{E}[\xi_i] &= \frac{\alpha}{\alpha + \beta} = \bar{\mu}_i, \\
\mathrm{Var}[\xi_i] &= \frac{\bar{\mu}_i(1 - \bar{\mu}_i)}{\kappa + 1}.
\end{align}
The mean exactly equals the curriculum trajectory, and the variance is inversely proportional to $\kappa + 1$. As $\kappa \to \infty$, the distribution concentrates at $\bar{\mu}_i$ (deterministic curriculum); as $\kappa \to 0$, it degenerates into independent uniform samples. In practice, $\kappa \in [3, 8]$ provides sufficient stochasticity to avoid geometry overfitting while keeping branch positions close to the curriculum.

\subsubsection{Discrete Mapping and Constraint Enforcement}

The continuous sample $\xi_i$ is mapped back to a discrete step index:
\begin{equation}
\tilde{s}_i = \left\lfloor s_{\min} + (s_{\max} - s_{\min}) \cdot \xi_i + 0.5 \right\rfloor.
\end{equation}
The final branch indices $\{s_1, s_2, s_3\}$ are obtained by sorting $\{\tilde{s}_1, \tilde{s}_2, \tilde{s}_3\}$.

\subsection{Trajectory Log-Probability}

Between branch points, the denoising trajectory follows deterministic ODE integration, which does not contribute to the trajectory log-probability (the transition is deterministic with Jacobian $1$). The trajectory log-probability therefore accumulates only the $T$ stochastic SDE transitions:
\begin{equation}
\log P_\theta(\tau) = \sum_{i=1}^{T} \log \pi_\theta(x_{s_i^-} \mid x_{s_i}),
\end{equation}
where each transition probability follows the Gaussian:
\begin{equation}
\log \pi_\theta(x_{s_i^-} \mid x_{s_i}) = -\frac{\|x_{s_i^-} - \mu_\theta(x_{s_i}, s_i)\|^2}{2\gamma_i^2} - \frac{d}{2}\log(2\pi\gamma_i^2),
\end{equation}
with $d$ being the latent dimension. In practice, this is computed as a per-dimension mean to maintain $O(1)$ magnitude regardless of latent resolution, which is essential for numerical stability in mixed-precision (bf16) training:
\begin{equation}
\overline{\log \pi}_\theta(x_{s_i^-} \mid x_{s_i}) \triangleq \frac{1}{d}\log \pi_\theta(x_{s_i^-} \mid x_{s_i}).
\end{equation}
This uniform scaling by $1/d$ is applied consistently to \emph{all} log-probability computations, including both the within-group Softmax normalization $p_i = \mathrm{softmax}_i(\sum_t \overline{\log\pi})$ and the IS ratio $\log\hat{w}_{i,t}$, so that the scaling cancels in any ratio or difference. Specifically: (i)~for the Softmax-TB advantage, the $1/d$ factor is identical across numerator and denominator of $p_i$ and thus cancels in $A_i = \log q_i - \log p_i$; (ii)~for the IS ratio, $\log\hat{w}_{i,t} = (\overline{\log\pi}_\theta - \overline{\log\pi}_{\text{old}}) \cdot d / d$ preserves the same clipping dynamics since both the current and old policies share the same scaling. The inverse temperature $\beta$ in the Boltzmann target operates on the reward scale and is unaffected by the log-probability scaling. Note that $\gamma_i$ is a constant determined by the noise schedule and does not depend on $\theta$; the gradient of $\log P_\theta(\tau)$ with respect to $\theta$ flows exclusively through the drift term $\mu_\theta$.

\section{Further Details on the Experimental Setup}\label{app:sec:exp-setup-details}

\subsection{Quality Metrics}\label{app:subsubsec:quality_metric}

We adopt the following evaluation metrics:

\begin{itemize}[leftmargin=15pt]
    \item \textbf{HPS-v2.1}~\cite{wu2023hpsv2}: A human preference score trained on large-scale pairwise human annotations, computed as the cosine similarity between image and text embeddings from a ViT-H/14 backbone.
    \item \textbf{ImageReward}~\cite{xu2023imagereward}: A reward model trained on human preference annotations over text-to-image generations, providing a scalar score that correlates with human aesthetic and fidelity judgments.
    \item \textbf{PickScore}~\cite{kirstain2023pick}: A CLIP-based~\cite{radford2021learning} preference model trained on the Pick-a-Pic dataset. We adopt the flow-scaled scoring mode during training.
    \item \textbf{GenEval}~\cite{ghosh2024geneval}: A compositional generation benchmark that evaluates whether generated images faithfully reflect the objects, attributes, and spatial relations specified in the prompt. Each image is scored by a Mask2Former object detector combined with CLIP attribute matching, served via an HTTP endpoint. During training, we use strict scoring (binary $1$ if all criteria are satisfied, $0$ otherwise); during evaluation, we report the fraction of compositional criteria satisfied.
    \item \textbf{OCR Accuracy ($1{-}\text{NED}$)}: For the text-rendering setting, we measure character-level accuracy as $1 - \text{NED}$, where NED denotes the normalized edit distance between the recognized text and the ground-truth target string.
    \item \textbf{LGMD (Log Geometric Mean Distance)}: A latent-space diversity metric defined in the main text (Section~\ref{sec:experiments}). LGMD computes the logarithm of the dimension-normalized geometric mean of pairwise Euclidean distances between flattened VAE latent features. Because the geometric mean is dominated by the smallest pairwise distances, LGMD is highly sensitive to near-duplicate samples; positive values indicate healthy diversity, while negative values signal mode collapse.
    \item \textbf{Cosine Diversity (Cos.\,Div.)}: A semantic diversity metric following GARDO~\cite{he2025gardo}. It computes the mean pairwise cosine distance in DINOv2~\cite{oquab2024dinov2} ViT-L/14 feature space: $\text{Cos.\,Div.} = \frac{2}{N(N{-}1)}\sum_{i<j}(1 - \cos(\psi(x_i), \psi(x_j)))$. While LGMD captures low-level structural duplicates, Cos.\,Div.\ captures semantic layout and texture differences.
\end{itemize}

\subsection{Model and Reward Specification}\label{app:subsubsec:synthetic-model}

Table~\ref{tab:model_links} lists the backbone model and reward models used in our experiments.

\begin{table}[ht]
\centering
\caption{Models used in our experiments and their sources.}
\label{tab:model_links}
\resizebox{\linewidth}{!}{
\begin{tabular}{ll}
\toprule
\textbf{Model} & \textbf{Link} \\
\midrule
FLUX.1-dev~(backbone) & \url{https://huggingface.co/black-forest-labs/FLUX.1-dev} \\
HPS-v2.1~\cite{wu2023hpsv2} & \url{https://github.com/tgxs002/HPSv2} \\
ImageReward~\cite{xu2023imagereward} & \url{https://huggingface.co/THUDM/ImageReward} \\
PickScore~\cite{kirstain2023pick} & \url{https://huggingface.co/yuvalkirstain/PickScore_v1} \\
\bottomrule
\end{tabular}
}
\end{table}

\textbf{Backbone.} We use FLUX.1-dev, a guidance-distilled rectified flow transformer~\cite{peebles2023scalable}. LoRA adapters are applied with rank $r{=}64$ and scaling factor $\alpha{=}128$ ($\alpha/r{=}2$) to all linear projections in each transformer block, including: (i)~attention Q/K/V and output projections for both the image stream and the context (text-conditioned) stream, and (ii)~the GEGLU gate and down projections in both the image and context feed-forward networks. All generation uses a classifier-free guidance~\cite{ho2022classifier} scale of $3.5$ and a resolution of $512{\times}512$. Training rollouts use $6$ denoising steps for efficiency; evaluation uses $28$ steps for full-quality generation.

\textbf{Reward normalization.} In the joint preference setting, HPS-v2.1, ImageReward, and PickScore are combined at equal weight. Before summation, each reward is independently z-score normalized within the $K{=}27$ trajectory group: $\tilde{R}_m = (R_m - \mu_m) / (\sigma_m + \epsilon)$, where $\mu_m$ and $\sigma_m$ are the within-group mean and standard deviation of reward $m$, and $\epsilon = 10^{-8}$ prevents division by zero. This normalization ensures that rewards with different scales contribute equally to the Softmax-TB advantage.

\subsection{Training Pipeline}\label{app:subsubsec:training_pipeline}

Each TMPO training iteration proceeds as follows:
\begin{enumerate}[leftmargin=15pt]
    \item \textbf{Prompt sampling.} Eight prompts are drawn uniformly from the training set, one per GPU.
    \item \textbf{Tree rollout.} On each GPU, the stochastic tree sampler generates $K{=}27$ terminal trajectories for its prompt via $T{=}3$ branch points with $B{=}3$ children each, yielding $8 \times 27 = 216$ images per iteration. Between branch points, deterministic ODE integration advances the latent. When the first branch point falls at step~$1$, the $B$ child trajectories are initialized from independent random seeds; the remaining two branch points inject CPS-type SDE noise~\cite{wang2025cps} with coefficient $\eta{=}0.7$. Branch positions are determined by the curriculum-guided Beta schedule (Appendix~\ref{appendix:beta_schedule}).
    \item \textbf{Reward evaluation.} Each terminal image $x_0^{(i)}$ is decoded and scored by the reward model(s). Rewards are z-score normalized in the joint setting.
    \item \textbf{Advantage computation.} The detached Softmax-TB advantage $A_i^{\bot}$ is computed per Eq.~\eqref{eq:advantage}.
    \item \textbf{IS ratio recomputation.} The current policy $\pi_\theta$ recomputes per-step log-probabilities at each branch point, yielding the bias-corrected IS ratio $\hat{\rho}_i$ per Eq.~\eqref{eq:ratio}.
    \item \textbf{Policy update.} The PPO-style clipped loss $\mathcal{L}_{\text{TMPO}}$ (Eq.~\eqref{eq:softmax_tb_loss}) plus a KL reference penalty is minimized via AdamW. The old policy $\pi_{\text{old}}$ is the frozen policy from the latest rollout; its log-probabilities are stored before the gradient update and remain fixed throughout the IS updates. EMA-smoothed parameters (decay $0.9$, update interval $8$) are maintained separately and used only for evaluation and checkpoint saving.
\end{enumerate}

\subsection{Hyperparameter Specification}\label{app:subsubsec:synthetic-hyperp}

Except for task-specific adjustments noted below, TMPO hyperparameters are fixed across all training protocols. We use AdamW ($\beta_1{=}0.9$, $\beta_2{=}0.999$) with weight decay $10^{-4}$ and a cosine annealing schedule ($\eta_{\min}{=}0.1\times\text{lr}$). The learning rate is $3\times10^{-5}$ for GenEval, OCR, and Joint Preference, and $5\times10^{-5}$ for PickScore. We train for 1{,}000 steps (GenEval), 250 steps (OCR), 500 steps (PickScore), and 500 steps (Joint). Gradient norms are clipped to $0.5$ except for PickScore ($0.3$). Training uses bf16 mixed precision throughout. Each prompt produces $K{=}27$ terminal trajectories via a three-level prefix-sharing tree with $B{=}3$ children per branch point ($T{=}3$ branch points). The training denoising horizon is $6$ steps and the evaluation horizon is $28$ steps. IS ratios are clipped with $\varepsilon{=}0.2$. The reward temperature $\beta$ warms up linearly from $0.8$ to $2.0$ over 150 steps (100 for OCR). EMA-smoothed parameters (decay $0.9$, update interval $8$) are maintained for evaluation and checkpoint saving. The tree sampler uses CPS-type SDE noise injection~\cite{wang2025cps} with noise coefficient $\eta{=}0.7$. Branch positions follow the curriculum-guided Beta schedule (Appendix~\ref{appendix:beta_schedule}) with early positions $(1,2,3)$, late positions $(1,3,5)$, and concentration $\kappa{=}6.0$. LoRA is applied with rank $r{=}64$ and $\alpha{=}128$.

\subsection{Baseline Hyperparameters}\label{app:subsubsec:baseline-hyperp}

For fair comparison, all baselines use the same backbone (FLUX.1-dev with LoRA $r{=}64$, $\alpha{=}128$), the same reward models, and the same prompt set. Wherever possible, we adopt the default settings from the original papers~\cite{liu2025flowgrpo,li2025mixgrpo,ding2026treegrpo,he2025gardo}. All baselines use $K{=}27$ trajectories per prompt to match TMPO's group size. Flow-GRPO, MixGRPO, TreeGRPO, and GARDO share a learning rate of $5\times10^{-5}$, gradient norm clipping of $0.3$, and IS clipping $\varepsilon{=}0.2$. The KL coefficient is $\beta_{\text{KL}}{=}0.03$ for Flow-GRPO, MixGRPO, and TreeGRPO, and $0.04$ for GARDO. All GRPO-based methods use z-score advantage normalization (TreeGRPO normalizes within the tree group). Flow-GRPO and GARDO sample $K$ independent trajectories per prompt; MixGRPO uses mixed ODE-SDE sampling; TreeGRPO uses prefix-sharing tree rollouts. GARDO additionally employs DINOv2-based advantage reweighting for diversity preservation.

\subsection{Compute Resources Specification}\label{app:subsubsec:synthetic-compute}

All experiments are conducted on $8$ GPUs using HuggingFace \texttt{accelerate} for distributed training with FSDP. Training uses bf16 mixed precision with TF32-enabled cuDNN kernels. Iteration times reported in the main text are wall-clock measurements averaged over the full training run. A single TMPO training run takes approximately 9.5 wall-clock hours (76 GPU-hours) on the PickScore only protocol (500 steps), 5.3 wall-clock hours (42 GPU-hours) on the OCR only protocol (250 steps), and 25.5 wall-clock hours (204 GPU-hours) on the GenEval only protocol (1{,}000 steps).

\section{Extended Experimental Results}\label{app:sec:extended-exp-results}

\subsection{Joint Preference Training}\label{app:subsec:joint}

Table~\ref{tab:comparison_joint} evaluates all methods under joint training with three preference rewards at equal weight. TMPO obtains the best HPS-v2.1 and diversity while remaining competitive on ImageReward and PickScore, indicating that distribution matching scales naturally to multi-objective settings without per-reward tuning.

\begin{table}[t]
  \centering
  \caption{Joint preference training on FLUX.1-dev. RL methods use HPS-v2.1, ImageReward, and PickScore at equal weight. Best values are \textbf{bold} and second-best are \underline{underlined}.}
  \label{tab:comparison_joint}
  \resizebox{\columnwidth}{!}{
    \begin{tabular}{lcccccc}
      \toprule
      \multirow{2}{*}{\textbf{Method}} & \multirow{2}{*}{\textbf{Iter. Time (s)} $\downarrow$} & \multicolumn{3}{c}{\textbf{Human Preference Alignment} $\uparrow$} & \multirow{2}{*}{\textbf{LGMD} $\uparrow$} & \multirow{2}{*}{\textbf{Cos.\,Div.} $\uparrow$} \\
      \cmidrule(lr){3-5}
      & & \textbf{HPS-v2.1} & \textbf{ImageReward} & \textbf{PickScore} & & \\
      \midrule
      \rowcolor{gray!10} FLUX.1-dev & --- & 0.310 & 1.119 & 22.604 & \underline{-0.056} & \underline{0.221} \\
      \midrule
      Flow-GRPO & 119.4 & 0.342 & \textbf{1.622} & 22.769 & -0.102 & 0.199 \\
      MixGRPO & 106.5 & 0.351 & 1.590 & 23.717 & -0.308 & 0.175 \\
      TreeGRPO & \underline{87.6} & \underline{0.355} & 1.603 & \textbf{23.951} & -0.282 & 0.178 \\
      GARDO & 125.9 & 0.341 & 1.574 & 22.927 & -0.184 & 0.209 \\
      \midrule
      \rowcolor{gray!20}
      \textbf{TMPO (Ours)} & \textbf{72.1} & \textbf{0.360} & \underline{1.613} & \underline{23.842} & \textbf{0.132} & \textbf{0.243} \\
      \bottomrule
    \end{tabular}
  }
\end{table}

\subsection{Generalization to SD3.5-Medium}\label{app:subsec:sd35}

To verify that TMPO generalizes beyond the FLUX.1-dev backbone, we evaluate all methods under the PickScore only protocol on SD3.5-Medium~\cite{esser2024sd3} with LoRA fine-tuning. All methods use the same LoRA rank and learning rate; all other hyperparameters follow their respective FLUX configurations.

\begin{table}[ht]
  \centering
  \caption{PickScore only training on SD3.5-Medium. All methods use LoRA fine-tuning. Best values are \textbf{bold} and second-best are \underline{underlined}.}
  \label{tab:sd35}
  \resizebox{\columnwidth}{!}{
  \begin{tabular}{lcccccc}
      \toprule
      \textbf{Method} & \textbf{Time (s)} $\downarrow$ & \textbf{HPS-v2.1} $\uparrow$ & \textbf{ImgRwd} $\uparrow$ & \textbf{PickScore} $\uparrow$ & \textbf{LGMD} $\uparrow$ & \textbf{Cos.\,Div.} $\uparrow$ \\
      \midrule
      \rowcolor{gray!10} SD3.5-Medium & --- & 0.272 & 0.893 & 21.592 & -0.061 & 0.213 \\
      Flow-GRPO & 101.8 & \underline{0.355} & 1.329 & \underline{23.289} & -0.123 & 0.195 \\
      MixGRPO & 96.2 & 0.348 & \underline{1.351} & 23.019 & -0.238 & 0.175 \\
      TreeGRPO & \underline{74.6} & 0.351 & 1.298 & 22.817 & -0.289 & 0.168 \\
      GARDO & 105.1 & 0.329 & 1.335 & 23.102 & \underline{0.072} & \underline{0.227} \\
      \rowcolor{gray!20}
      \textbf{TMPO (Ours)} & \textbf{63.9} & \textbf{0.361} & \textbf{1.358} & \textbf{23.551} & \textbf{0.118} & \textbf{0.239} \\
      \bottomrule
  \end{tabular}
  }
\end{table}

The results mirror the FLUX.1-dev findings: TMPO achieves the best scores on all metrics with a consistent ${\sim}37\%$ iteration-time reduction over Flow-GRPO. Flow-GRPO, MixGRPO, and TreeGRPO again exhibit negative LGMD; GARDO preserves positive LGMD through its explicit diversity mechanism but still trails TMPO on both diversity and reward metrics, confirming that Softmax-TB provides a stronger diversity guarantee without auxiliary reweighting.

\subsection{Qualitative Comparison with Baselines}\label{app:subsec:qualitative_results}

\begin{figure}[!htbp]
  \centering
  \includegraphics[width=\textwidth]{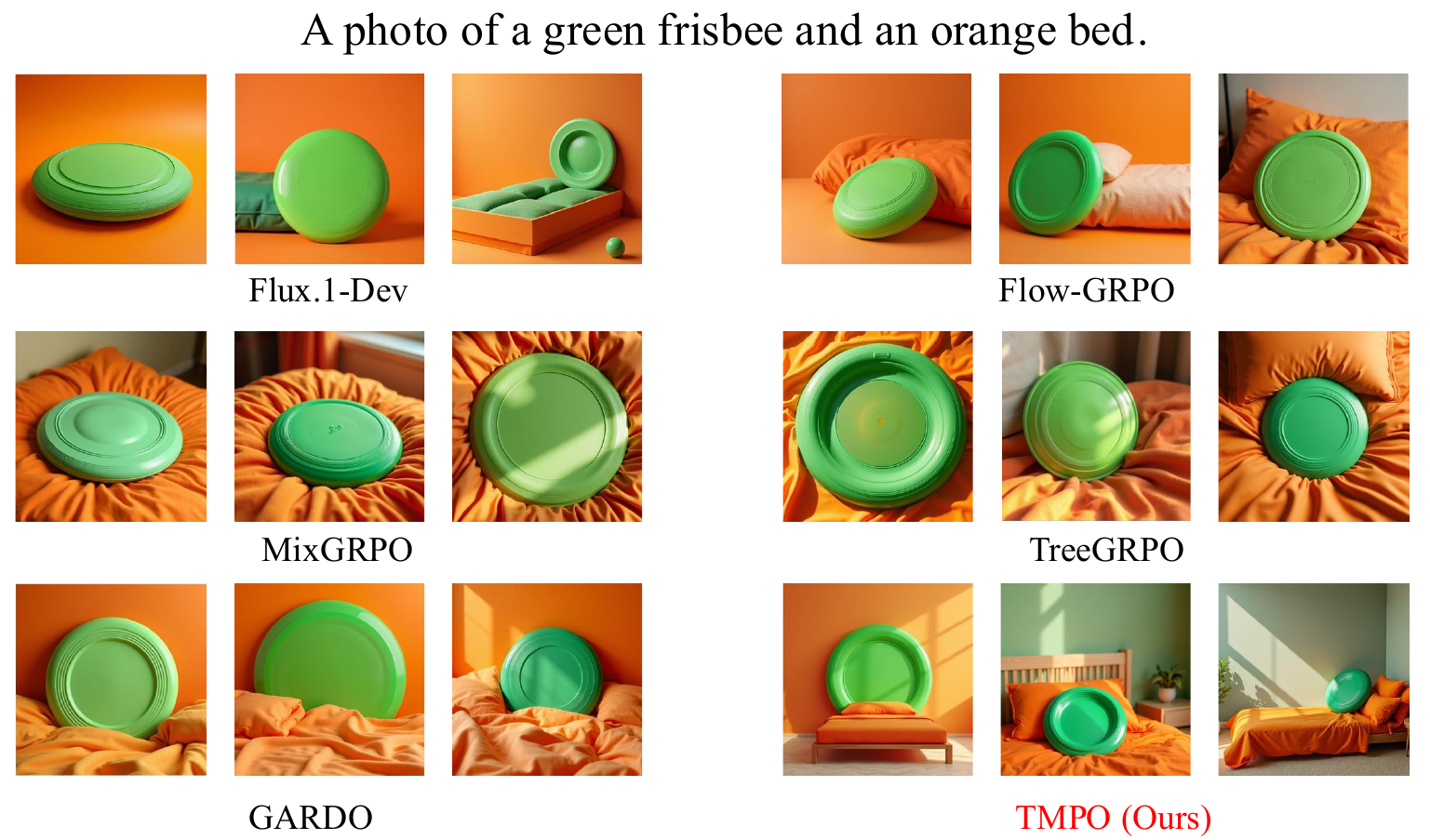}
  \caption{\textbf{Qualitative comparison on GenEval (compositional image generation).} TMPO faithfully renders all specified objects, attributes, and spatial relations while maintaining diverse compositions across samples.}
  \label{fig:qual_geneval}
\end{figure}

\begin{figure}[!htbp]
  \centering
  \includegraphics[width=\textwidth]{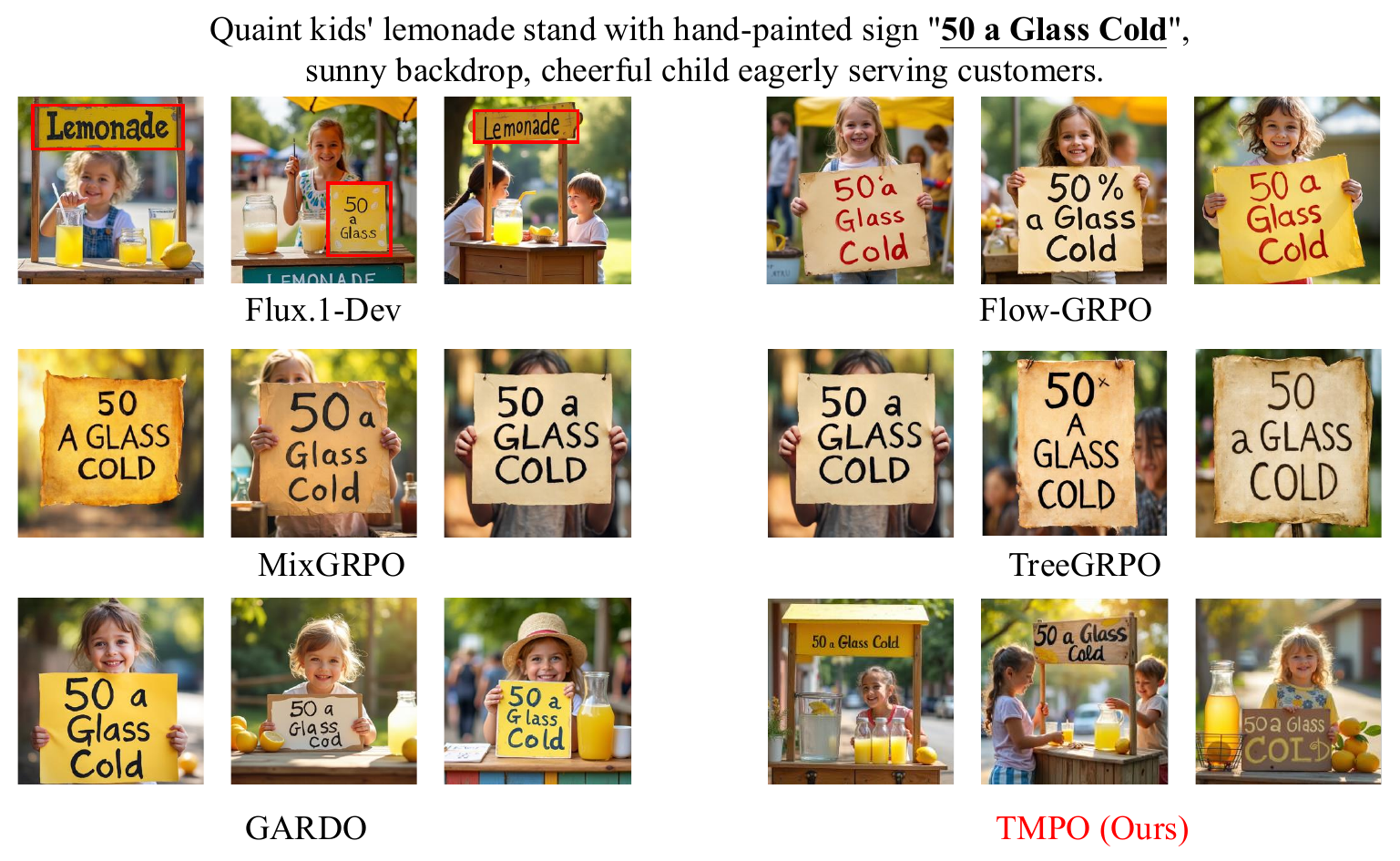}
  \caption{\textbf{Qualitative comparison on OCR (visual text rendering).} TMPO accurately renders the target text strings with high legibility while preserving visual diversity in background and style, whereas baselines either produce near-duplicate layouts or exhibit text rendering errors.}
  \label{fig:qual_ocr}
\end{figure}

\begin{figure}[!htbp]
  \centering
  \includegraphics[width=\textwidth]{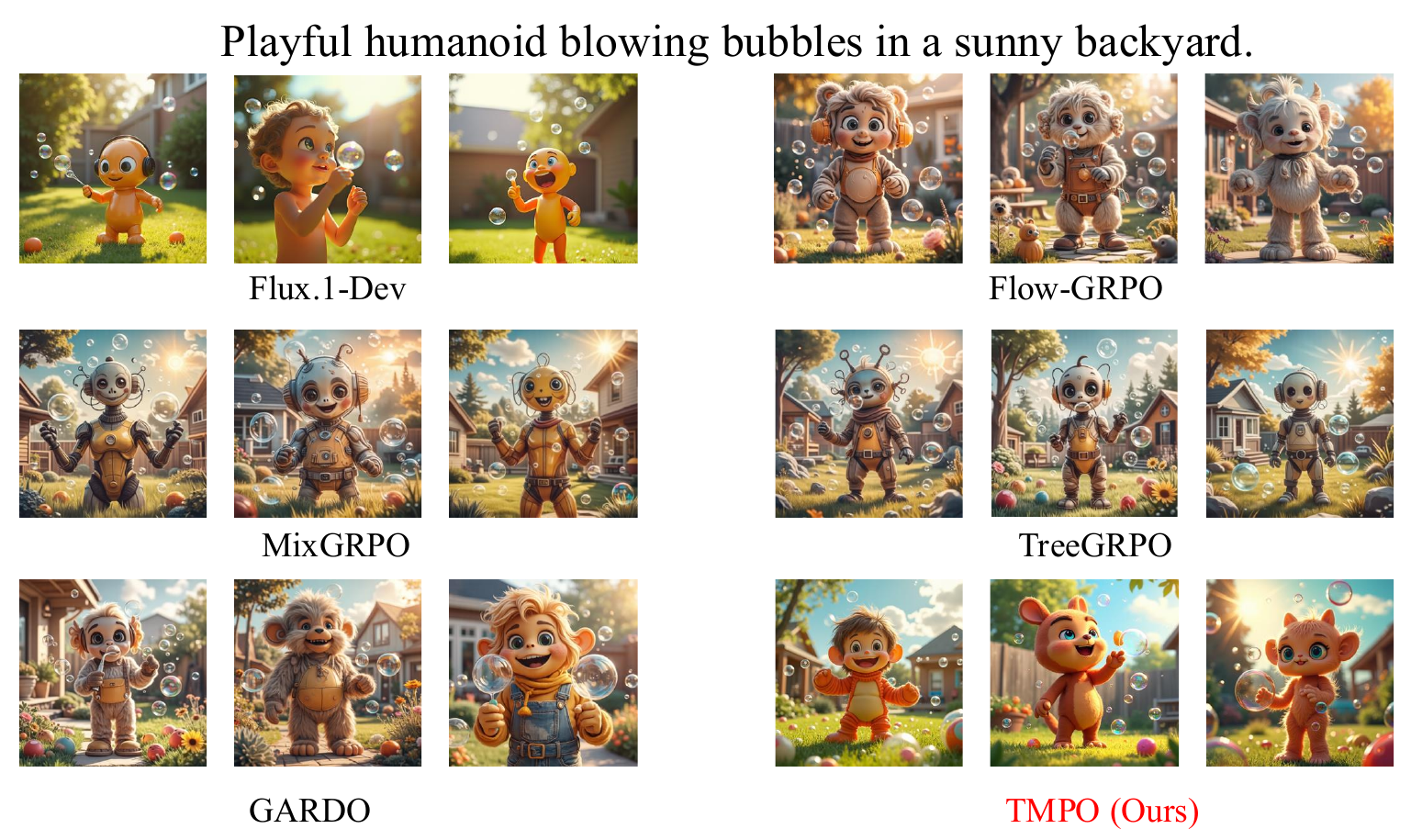}
  \caption{\textbf{Qualitative comparison on PickScore (human preference alignment).} TMPO produces aesthetically appealing and prompt-faithful images with noticeably greater diversity in composition, color palette, and viewpoint compared to baselines.}
  \label{fig:qual_pick}
\end{figure}

Figures~\ref{fig:qual_geneval},~\ref{fig:qual_ocr}, and~\ref{fig:qual_pick} present qualitative comparisons between TMPO and baseline methods on the GenEval (compositional image generation), OCR (visual text rendering), and PickScore (human preference alignment) protocols, respectively. Across all three tasks, TMPO generates images that are both faithful to the text prompt and visually diverse, whereas GRPO-based baselines tend to produce near-duplicate outputs or sacrifice prompt fidelity for reward maximization.

\subsection{Evolution of Evaluation Images Across Training Steps}\label{app:subsec:training_evolution}

\begin{figure}[!htbp]
  \centering
  \includegraphics[width=\textwidth]{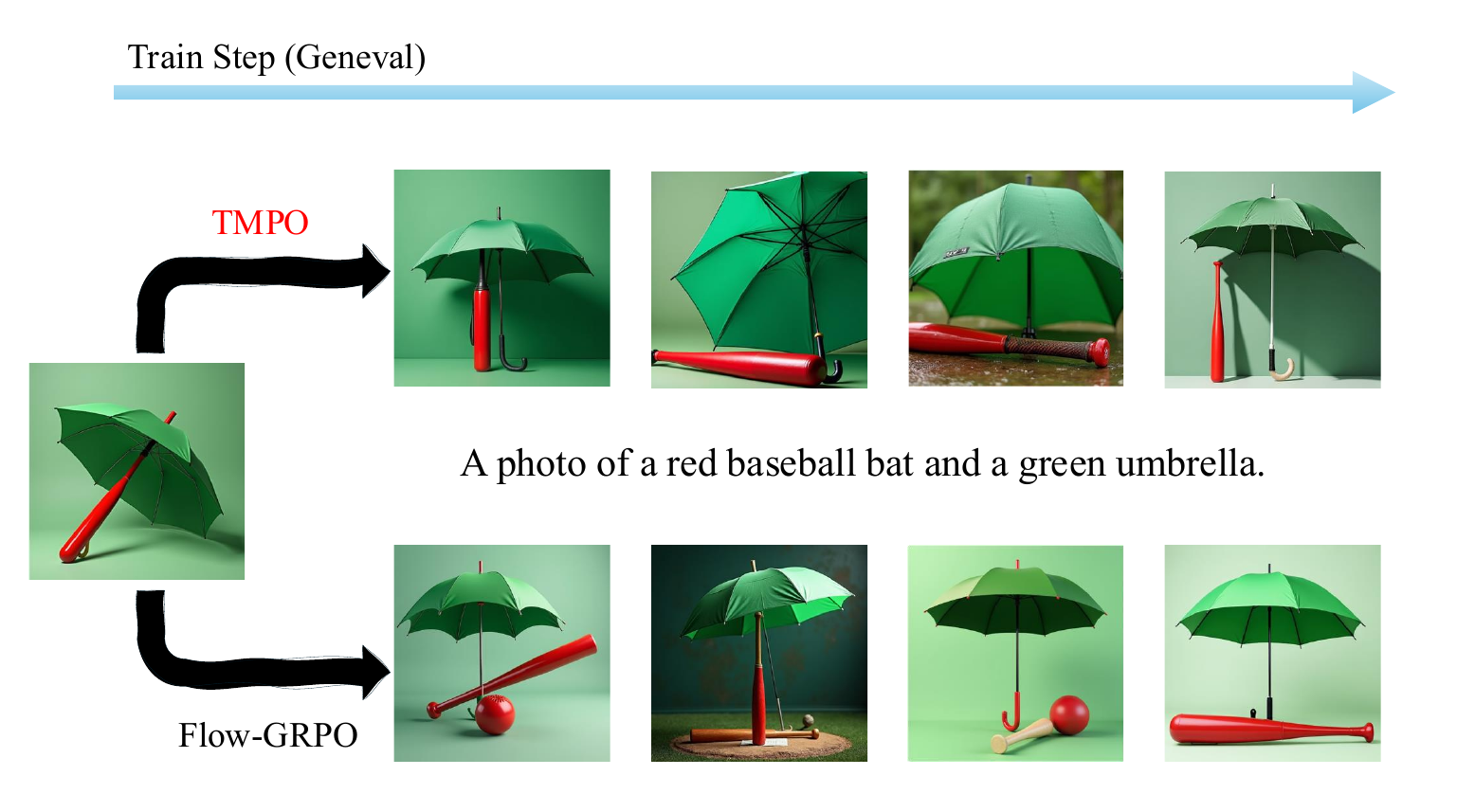}
  \caption{\textbf{Evolution of generated images across training steps on GenEval.} TMPO progressively improves compositional accuracy while maintaining diverse object arrangements and visual styles throughout training.}
  \label{fig:evo_geneval}
\end{figure}

\begin{figure}[!htbp]
  \centering
  \includegraphics[width=\textwidth]{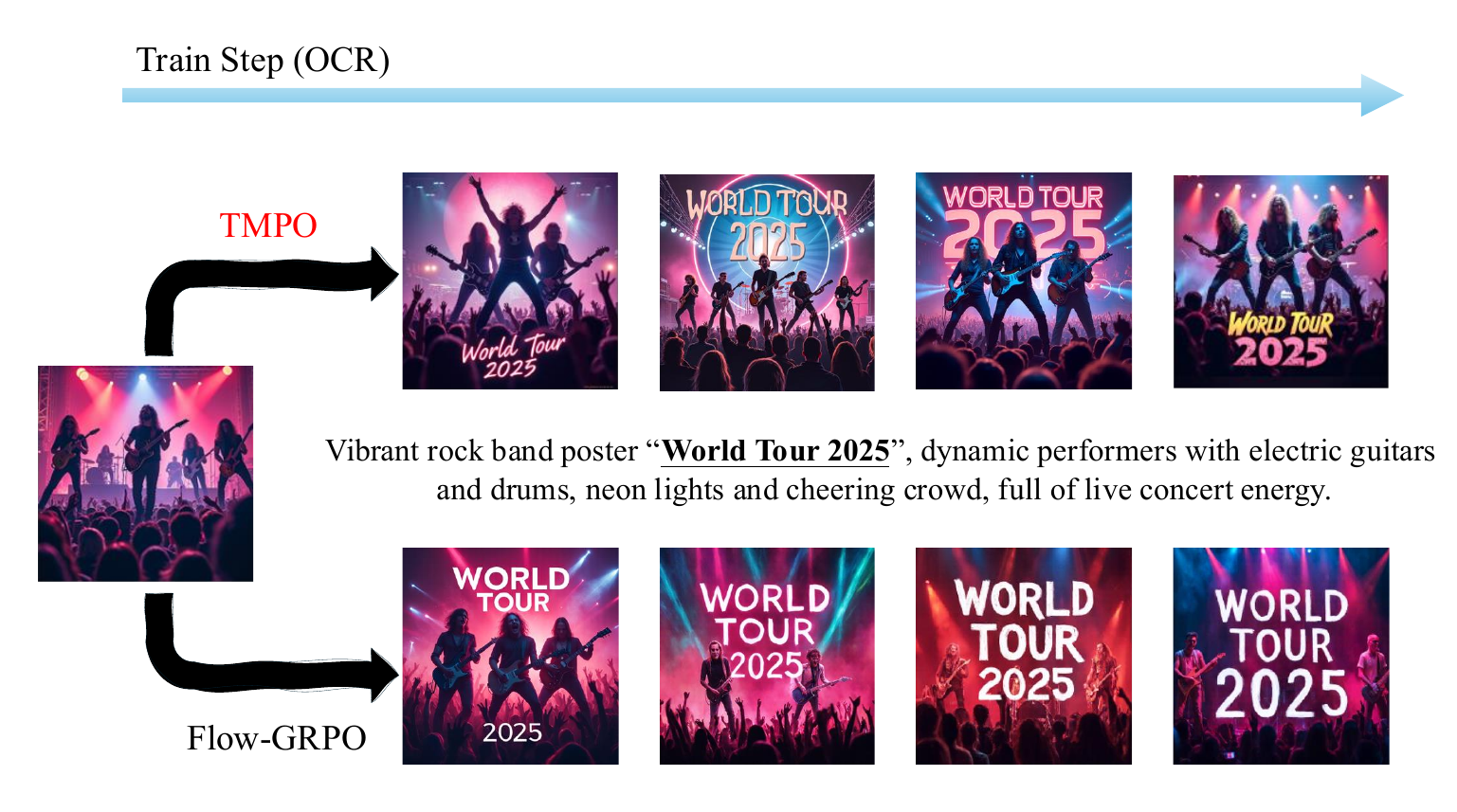}
  \caption{\textbf{Evolution of generated images across training steps on OCR.} Text rendering quality improves steadily, with the model learning to produce legible characters while retaining diverse visual layouts.}
  \label{fig:evo_ocr}
\end{figure}

\begin{figure}[!htbp]
  \centering
  \includegraphics[width=\textwidth]{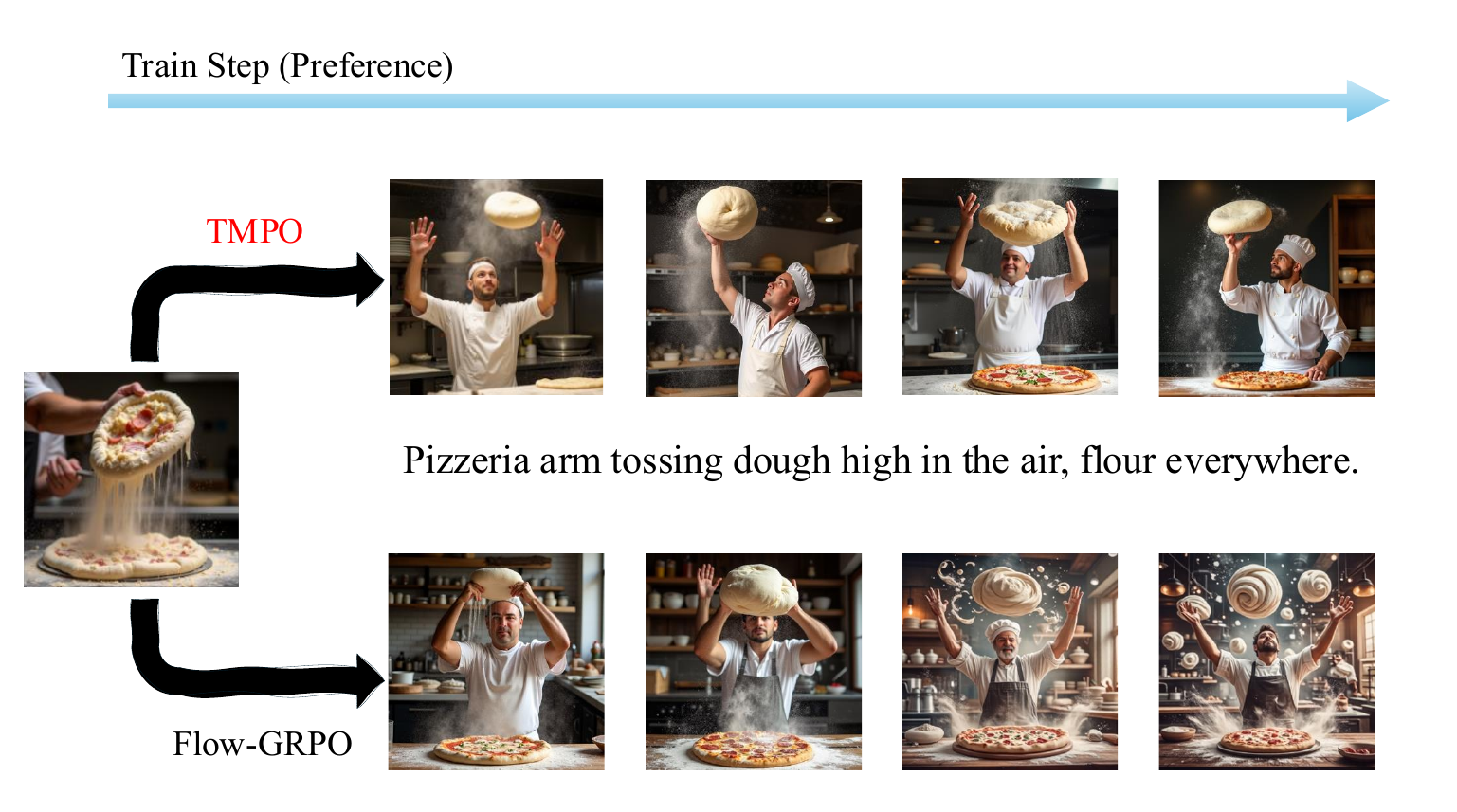}
  \caption{\textbf{Evolution of generated images across training steps on PickScore.} Image aesthetics and prompt fidelity improve progressively, while sample diversity is preserved even at later training stages.}
  \label{fig:evo_pick}
\end{figure}

Figures~\ref{fig:evo_geneval},~\ref{fig:evo_ocr}, and~\ref{fig:evo_pick} visualize how generated images evolve across training steps under the GenEval, OCR, and PickScore protocols, respectively. As training progresses, TMPO steadily improves task-specific quality (compositional correctness, text legibility, and aesthetic appeal) while preserving sample diversity throughout the optimization process, avoiding the mode collapse commonly observed in reward-maximizing methods.

\end{document}